**Mapping smallholder cashew plantations to inform sustainable tree crop expansion in Benin**

Leikun Yin[1], Rahul Ghosh[2], Chenxi Lin[1], David Hale[3], Christoph Weigl[3,4], James Obarowski[5], Junxiong Zhou[1], Jessica Till[1], Xiaowei Jia[6], Troy Mao[7], Vipin Kumar[2], Zhenong Jin[1*]

1 Department of Bioproducts and Biosystems Engineering, University of Minnesota, St. Paul, MN 55108, USA

2 Department of Computer Science and Engineering, University of Minnesota, Minneapolis, MN 55414, USA

3 TechnoServe Labs, Arlington, VA 22209, USA

4 McKinsey & Company, New York City, NY 10022, USA

5 PULA, Nairobi, Nairobi County 00800, Kenya

6 Department of Computer Science, University of Pittsburgh, Pittsburgh, PA 15260, USA

7 College of Letters and Science, University of California, Davis, CA 95616, USA

Corresponding authors*: jinzn@umn.edu (Jin Z.)

**Abstract**

Cashews are grown by over 3 million smallholder farmers in more than 40 countries worldwide as a principal source of income. Expanding the area of cashew plantations and increasing productivity are critical to improving the livelihood of many smallholder communities. As the third largest cashew producer in Africa, Benin has nearly 200,000 smallholder cashew growers contributing 15% of the country's national export earnings. Expansion of the cashew industry is thus an essential economic driver and a governmental priority in Benin. However, a lack of information on where and how cashew trees grow across the country hinders decision-making that could support increased cashew production and poverty alleviation. By leveraging 2.4-m Planet Basemaps and 0.5-m aerial imagery, two newly developed deep learning algorithms, and large-scale ground truth datasets, we successfully produced the first-of-its-kind national map of cashew in Benin and characterized the expansion of cashew plantations between 2015 and 2021. In particular, we developed a SpatioTemporal Classification with Attention (STCA) model to map the distribution of cashew plantations, which can fully capture texture information from discriminative time steps during a growing season. We further developed a Clustering Augmented Self-supervised Temporal Classification (CASTC) model to distinguish high-density versus low-density cashew plantations by automatic feature extraction and optimized clustering. Results show that the STCA model has an overall accuracy over 85% based on 1,400 ground truth point samples from each year. The CASTC model achieved an overall accuracy of 76% based on 348 ground truth samples of planting density. We found that the cashew area in Benin has almost doubled to 519 ± 20 kha from 2015 to 2021 with 60% of new plantation development coming from cropland or fallow land, while encroachment of cashew plantations into protected areas has increased by 55%. Only about half of cashew plantations were high-density in 2021, suggesting high potential for intensification. Our study illustrates the power of combining high-resolution remote sensing imagery and state-of-the-art deep learning algorithms to better understand tree crops in the heterogeneous smallholder landscape, which can help efficiently allocate limited training and nursery resources for sustainable agricultural development.

# 1 Introduction

Achieving zero hunger and ending poverty by 2030 are two primary and vital missions of the United Nations (UN) Sustainable Development Goals (UN, 2015). Poverty is a leading cause of hunger, which continues to affect many people in food-insecure regions of South Asia and Sub-Saharan Africa (Lowder et al., 2016; Samberg et al., 2016). Most research and applications in the field of food security focus on increasing staple crop production in the developing world, e.g., through crop yield forecasts (Basso et al., 2019) and early warning of famine situations (Becker-Reshef et al., 2020). However, these scientific explorations cannot resolve the underlying problem of poverty for countries in the Global South. For smallholder farmers, growing tree crops can provide a stable source of income because of their relatively high cash value, predictable yields, long tree lifespans of 20-30 years, and good adaptability to growth conditions (Lin et al., 2021). With tens of millions of smallholder farmers in tree crop production (cashew, cocoa, coffee, etc.), tree crops contribute a large percentage of income in poor communities globally (Waarts et al., 2021). Increasing the production of tree crops is therefore an important and effective way to improve the living conditions of smallholders.

Cashew tree crops are widely farmed in 46 countries across Africa, Asia, Latin America, and the Caribbean, 18 of which are among the least developed countries. Africa accounts for more than half of global raw cashew nut production followed by Asia, and ~80% of Africa's raw cashew nut production is concentrated in West Africa, principally in Côte d'Ivoire, Nigeria, Benin, and Guinea-Bissau. However, less than 10% of the world's raw cashew nuts are processed in Africa, and the majority (more than 85%) are processed in Asia, primarily in Vietnam and India (UNCTAD, 2021). In addition to cashew nuts, the by-products of the cashew crop have a variety of industrial uses that can assist smallholders diversify their sources of income. Cashew nut shell liquid has the potential to be used as a biofuel (Sanjeeva et al., 2014), and cashew apples can be used to produce a range of beverages and animal feed (Gomes et al., 2018). Since cashew tree crops are typically cultivated by smallholder farmers, value additions in the cashew industry and poverty reduction are tightly connected.

Benin is among the top ten cashew growers in the world and the third largest cashew nut producer in West Africa (Duguma et al., 2021). Cashew nuts from Benin are renowned for their superior quality and bright white hue. The government of Benin recognizes the importance of cashew tree plantations in the fight against poverty for smallholder farmers, as evidenced by their strategy to double cashew production from 2016 to 2021 in the government action plan - Benin Revealed: Government Action Program (MAEP-Benin, 2017) - and the subsequent 2022-2026 plan (PNIASAN-Benin, 2022). There are two main ways to increase cashew nut production. The first strategy is to expand the cashew plantation area under cultivation by converting other land use types. The second is to improve the use of good agricultural practices (GAP) on existing cashew plantations to increase yield. The two strategies require an understanding of cashew plantation distribution, i.e., detailed knowledge regarding the location of cashew growing areas around the country and GAP implementation including planting density (tree spacing). Knowledge of the spatial distribution of cashew plantations and planting density is therefore essential to inform government policies regarding land use conversion and to efficiently manage field extension services. Periodic mapping can help the government understand how policies have been implemented and update land-use guidelines. In addition, for GAP related to planting density, identifying regions where cashews have been planted with below optimal GAP planting density guidelines can help best direct needed resources, including the provisioning of new cashew nurseries that are a core part of Benin's national cashew strategy and are crucial to increasing cashew production. In Benin and many other countries in sub-Saharan Africa, the traditional method of gathering information about cashew plantation distribution and GAP related to tree density is through extensive field data collection by conducting in-person farm visits and field surveys, which are inefficient and laborious on a large scale. Efficient and low-cost measures are urgently needed to assist the implementation of development programs on a national scale (Burke and Lobell, 2017; Chivasa et al., 2017).

Remote sensing offers a cost-effective and time-saving way to characterize ground objects and track surface changes over time at a large scale. Agricultural remote sensing has made significant contributions to fighting food insecurity through mapping crop types, predicting yields, and monitoring crop diseases/insect pests and more (Atzberger 2013; Maes and Steppe 2019; Nellis et al., 2009; Sishodia et al., 2020). However, research involving remote sensing and tree crops is mostly concerned with mapping tree crops and the effects of deforestation, with an emphasis on oil palm (Cheng et al., 2016; Gutiérrez-Vélez and DeFries, 2013; Xu et al., 2022) and rubber (Dong et al., 2013; Tridawati et al., 2020) trees. Rubber and oil palm trees have relatively prominent spatial and spectral features, which reduces the need for high spatial resolution imagery. For example, the crown diameter of a mature oil palm tree can be ~10 m and they are often grown in commercial farmlands spanning a few kilometers (Lin et al., 2021). In contrast, certain essential global commodity tree crops, such as cashew trees, have smaller crowns and are planted in irregular patterns within small plots (< 5 ha). In general, cashew plantations have received little attention in the past, with the exception of some local-scale studies in a specific community or national park with medium resolution (10-30 m) remote sensing imagery and traditional machine learning methods (Pereira et al., 2022; Singh et al., 2018). Those approaches are not transferable to national-scale studies and are insufficient to resolve individual fields that are particularly relevant to smallholder management practices. Furthermore, mapping smallholder tree crops remains challenging given the fragmentary landscape and limited satellite observation capabilities. Although a few studies have targeted smallholder plantation systems and complex landscapes (Ballester-Berman and Rastoll-Gimenez 2021; Descals et al., 2019; Dong et al., 2012; Maskell et al., 2021), the observational ability of medium-resolution spatiotemporal satellite imagery (e.g., MODIS, Landsat, and Sentinel) is limited for small-crown tree crops. Even fewer studies address different intra-class management practices, e.g., coffee sub-categories (Hunt et al., 2020; Kawakubo and Pérez Machado, 2016; Maskell et al., 2021). In recent years, in addition to aerial and drone imagery, as a number of microsatellite constellations have been launched by private aerospace companies such as Planet, Airbus, and DigitalGlobe, more data sources are available for high-resolution mapping (< 3 m). However, only limited small-scale case studies of ~1000 $km^2$ or less have leveraged high-resolution data to map smallholder tree crops (Burnett et al., 2019; Cui et al., 2022).

To understand smallholder cashew plantations in Benin and fill the gap in large-scale small-crown tree crop mapping with high-resolution imagery, we employed Planet Basemaps (2.4 m) and aerial imagery (0.5 m) with advanced spatiotemporal deep learning techniques to map cashew plantation distribution and planting density in Benin. Several aspects of smallholder cashew plantations made the two classification tasks challenging. First, cashew trees of different varieties and ages grow in the same plantation with irregular spacing, which results in tremendous intra-plantation heterogeneity. Second, there is often large inter-plantation heterogeneity in plantation size, shape, and planting density. In this case, sensors with medium spatial resolution (e.g., Sentinel and Landsat) and traditional machine learning classification algorithms are not sufficient for mapping cashew plantations at the field level. Finer remote sensing products (such as the 2.4-m

Planet Basemaps and 0.5-m aerial imagery used here) and advanced deep learning techniques have opened up new possibilities for monitoring smallholder cashew plantations. Third, omnipresent clouds and shadows distort the spectral signal captured by sensors. The daily revisit frequency of Planet imagery helps ensure the availability of cloud-free Planet Basemaps on a monthly basis.

To our knowledge, this is the first-of-its-kind large-scale cashew plantation map leveraging high-resolution remote sensing imagery. In this study, we (i) developed spatiotemporal tree classification algorithms for cashew, (ii) mapped cashew plantation spatial distributions for four years (2015, 2019, 2020, and 2021), (iii) tracked the spatiotemporal changes in cashew plantations, and (iv) distinguished high- and low-density cashew plantation management practices. While this study focuses on cashew tree crops in Benin as a special case, the derived classification algorithm can help inform cashew mapping in other countries or be used for other similar tree crops.

## 2 Data and methods

### 2.1 Study area

The Republic of Benin comprises 12 departments (the primary administrative units) and is subdivided into 77 communes. The study area located in central Benin (1-3°E, 7-10°N) is one of the primary cashew-growing regions in West Africa, which spans 12 communes in four departments - Donga, Borgou, Collines, and Zou. (Fig. 1(a)). Heterogeneous landscapes are prevalent here (Fig. 1(b)), and cashew trees are typically cultivated in smallholder plantations of less than 5 ha.

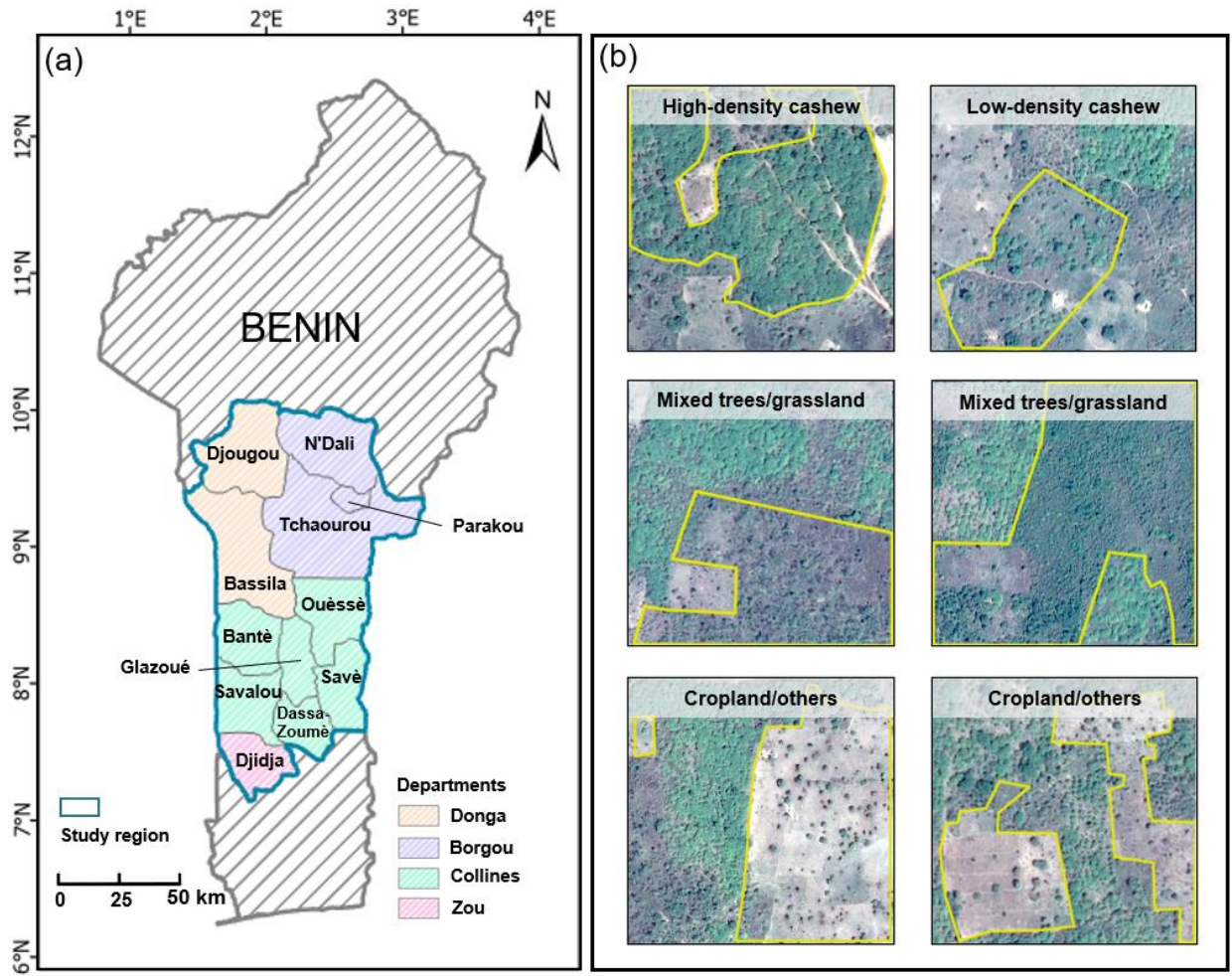

Fig. 1. (a) Location of the study region and administration map. (b) Sample cashew plantation, mixed trees/grassland, and cropland/others. The extents enclosed by the yellow boundaries correspond to the categories noted.

The study region has a tropical savanna climate, and its typical yearly temperature ranges from 24 to 31 °C. In the study region, the dry season lasts from November to April, while the rest of the year makes up the rainy season (Table 1). Drought conditions are concentrated between December and January during the dry season, while the wettest months occur from June to September during the rainy season. The planting time for cashew tree seedlings is during the rainy season, mainly between July and August. The cashew tree blossoms and produces fruit mainly during the dry season. The peak flowering period for the cashew tree lasts from December through January and cashew nut harvest typically takes place from February through March.

Table 1. Average precipitation in millimeters per month from 2015 to 2021 (CHIRPS) and cashew crop calendar in which dark colors indicate more concentrated events.

| | | May | June | July | Aug. | Sept. | Oct. | Nov. | Dec. | Jan. | Feb. | Mar. | Apr. |
|---|---|---|---|---|---|---|---|---|---|---|---|---|---|
| Precipitation (mm/month) | Rainy | 134 | 164 | 198 | 205 | 203 | 114 | | | | | | |
| | Dry | | | | | | | 11 | 4 | 3 | 11 | 59 | 75 |
| Cashew crop calendar | Planting | | | | | | | | | | | | |
| | Flowering | | | | | | | | | | | | |
| | Harvest | | | | | | | | | | | | |

## 2.2 Imagery

### 2.2.1 Planet Basemaps

Because the study region is located in the tropics, the image quality from optical sensors with low revisit frequencies is impaired significantly by year-round clouds and shadows. Furthermore, high spatial resolution is needed for mapping tasks because of the prevalence of irregular smallholder farms of less than 5 ha in the study region. For these reasons, we employed the monthly Planet Basemaps product to map cashew trees during less cloudy months (November to May) each year (2019-2021). Planet Basemaps is a 4-band (blue, green, red, and near infrared) surface reflectance (SR) product composed of images captured by the PlanetScope microsatellite constellation (Fig. S1(b)-(c)). The PlanetScope consists of hundreds of microsatellites carrying three generations of sensors (Dove Classic, Dove-R, and SuperDove), and the raw imagery it captures has a resolution of about 3 m spatially and 1 day temporally (Planet, 2022a). To develop the Planet Basemaps, raw imagery has undergone radiometric correction, geometric correction, orthorectification, and quality imagery selection. To reduce the variance across images and mitigate the effects of the atmosphere, the Planet Basemaps have also been standardized to a mature MODIS

SR product (Planet, 2022b). The Planet Basemaps are organized in tiles, with 569 tiles required each month to fully cover the study region. Two levels of the Planet Basemaps are available: Zoom Level 15 (4.77 m) and Zoom Level 16 (2.38 m), and we opted for the latter given its better observation capability.

### 2.2.2 Aerial imagery

Aerial imagery was employed for cashew plantation mapping in 2015 due to the fact that Planet Basemaps are not available prior to 2019. The aerial imagery came from a project supervised by the Benin government for the preservation and development of gallery forests and digital base mapping production. A fixed wing aircraft, a Piper PA-31 Navajo, was used for imagery collection in May 2015. A multispectral camera (UltraCam Eagle Mark 3) with a focal length of 40 mm was mounted on the airplane platform to collect ground information in four spectral bands (blue, green, red, near infrared) from 3 kilometers above the ground (Fig. S1(a)-(b)). Radiometric correction, geometric correction, and orthorectification have been applied to develop a SR product that is organized in grids. The SR product has a spatial resolution of 0.5 m, and 2031 tiles were required to fully cover the study region.

## 2.3 Ground survey truth

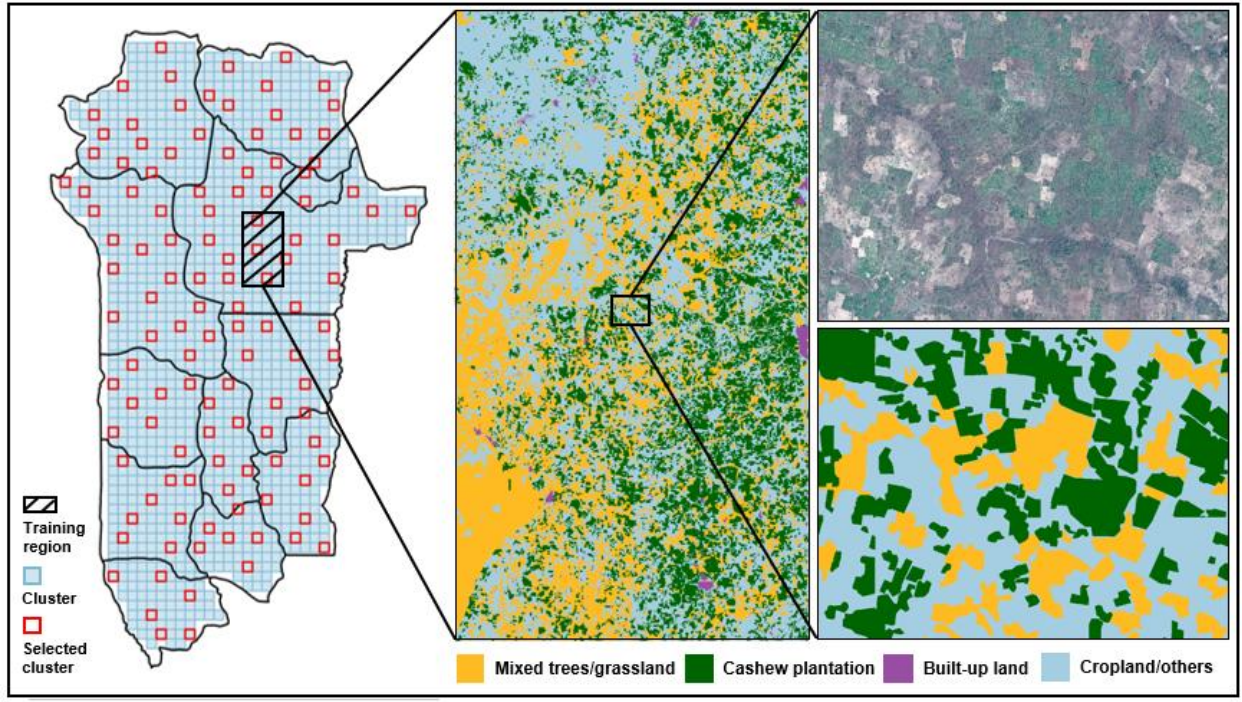

Fig. 2. Locations of the training region and validation data of selected clusters. Right-hand detailed panels show Airbus RGB imagery with manually delineated labels for comparison.

### 2.3.1 Training data

Our chosen region for training the classification algorithms is an area of 1,000 km$^2$ (Fig. 2), where the heterogeneous landscapes and irregular smallholder cashew plantations are appropriate for training and selecting the optimal deep learning model. For this training region, we launched a ground survey through the TechnoServe BeninCajù program to collect detailed land cover types, and aggregated them into four categories of cashew plantations, mixed trees/grassland, built-up land, and cropland/others. Based on this survey map, we manually delineated ground truth polygons of the four categories using 0.5-m aerial imagery in 2015 and Airbus imagery in 2020 for mapping cashew plantation distributions in 2015 and from 2019 to 2021, respectively. The training region in Fig. 2 shows labeled examples from Airbus imagery. The mixed trees/grassland class includes mixed scattered trees, grassland, and gallery forest, while the cropland/others class mainly includes cropland and bare land. Since there is some mismatch when the training labels based on 0.5-m imagery are directly applied to the 2.4-m Planet Basemaps, a resampling was applied. Because resampling from the manual ground truth to the Planet-based ground truth would cause a mixture of boundary pixels between two land cover types, we performed a 2-pixel erosion for each class, then relabeled eroded pixels as the cropland/others class and removed connected pixel clusters of less than 30 pixels.

For mapping cashew planting density, we defined high-density and low-density cashew plantations as having greater than or less than 100 trees/ha, respectively. In production practice, optimal yield is achieved in high-density plantations with a planting density between 100 and 180 cashew trees/ha. Low-density planting (< 100 trees/ha) cannot fully exploit the productive potential of cropland, while a density of over 180 trees/ha can lead to declining marginal productivity due to quality issues caused by disease and infestation. Although here we used a threshold of 100 trees/ha to distinguish between low-density and high-density cashew plantations, future studies will use a more in-depth classification that includes "very-high-density" plantations above 180 trees/ha.

### 2.3.2 Sampling strategy and validation

Our sampling strategy follows a probability sampling protocol (Olofsson et al., 2014) and consists of two phases to generate unbiased area estimates with uncertainties. In the first phase, we performed a simple random sampling method at the cluster level. The study region was divided into 5 by 5 km clusters totaling 1,663 clusters (Fig. 2), where the 5-km cluster size ensured the inclusion of a large number of smallholder fields. Of these, we discarded 241 clusters that were located on the study area boundary. The remaining 1422 clusters were retained for probability sampling, from which 120 sample clusters were randomly selected. In the second phase, we implemented a stratified random sampling method at the pixel level in the selected clusters. This sampling strategy significantly reduces the workload of visual interpretation by constraining sample pixels into these clusters. Specifically, according to our three mapped classes (mixed trees/grassland, cashew plantations, and cropland/others) we first generated four land-use change classes between successive years to estimate the area of source classes that account for cashew plantation expansion. Then, we constructed seven strata consisting of the three stable land cover classes and the four land cover change classes (listed below). We allocated the sample size in proportion to strata area, but slightly increased the sample size for less frequent classes, i.e., land-use change classes. This approach helps to balance the trade-off between user's, producer's, and overall accuracies (Olofsson et al., 2014). The seven strata of stable mixed trees/grassland, stable cashew plantations, stable cropland/others, change from mixed trees/grassland to cashew, change from cropland/others to cashew, change from mixed trees/grassland to cropland/others, and change from cropland/others to mixed trees/grassland contain 300, 200, 400, 100, 100, 100, and 200 sample points, respectively. A total of 1,400 sample pixels were selected for visual interpretation for each year. Five types of auxiliary information were additionally displayed to help visually interpret sample pixels: (1) 2.4-m temporal Planet Basemaps from 2019-2021, (2) 0.5-m aerial imagery for 2015, (3) high-resolution imagery from Google Earth Pro, (4) 11,397 field-collected cashew plantation boundaries, and (5) more than 1,100 field-collected and visually interpreted samples labeled by TechnoServe. The reason why we didn't use the field samples directly was that they were collected before this study and did not consider the need for unbiased accuracy and area estimation. In accordance with prior studies (Olofsson et al., 2013; Olofsson et al., 2014; Stehman 2014), we adjusted the accuracy and area estimation by considering the area of each stratum and generated 95% confidence intervals.

Furthermore, as the cashew tree is a perennial plant, we used 196 cashew plantation ground truth samples with the same exact location across the four years to validate the consistency of our classification maps in each of the four years. We also used 348 field-collected samples with known planting densities to assess the accuracy of cashew plantation planting density determinations. Prior to the ground survey, we selected validation point locations that allowed for both transportation accessibility and widespread distribution of the samples. Enumerators then used GPS devices to locate the predetermined samples and noted the planting density.

**2.4 Methodology**

**2.4.1 Tree crop mapping algorithms for cashew plantations**

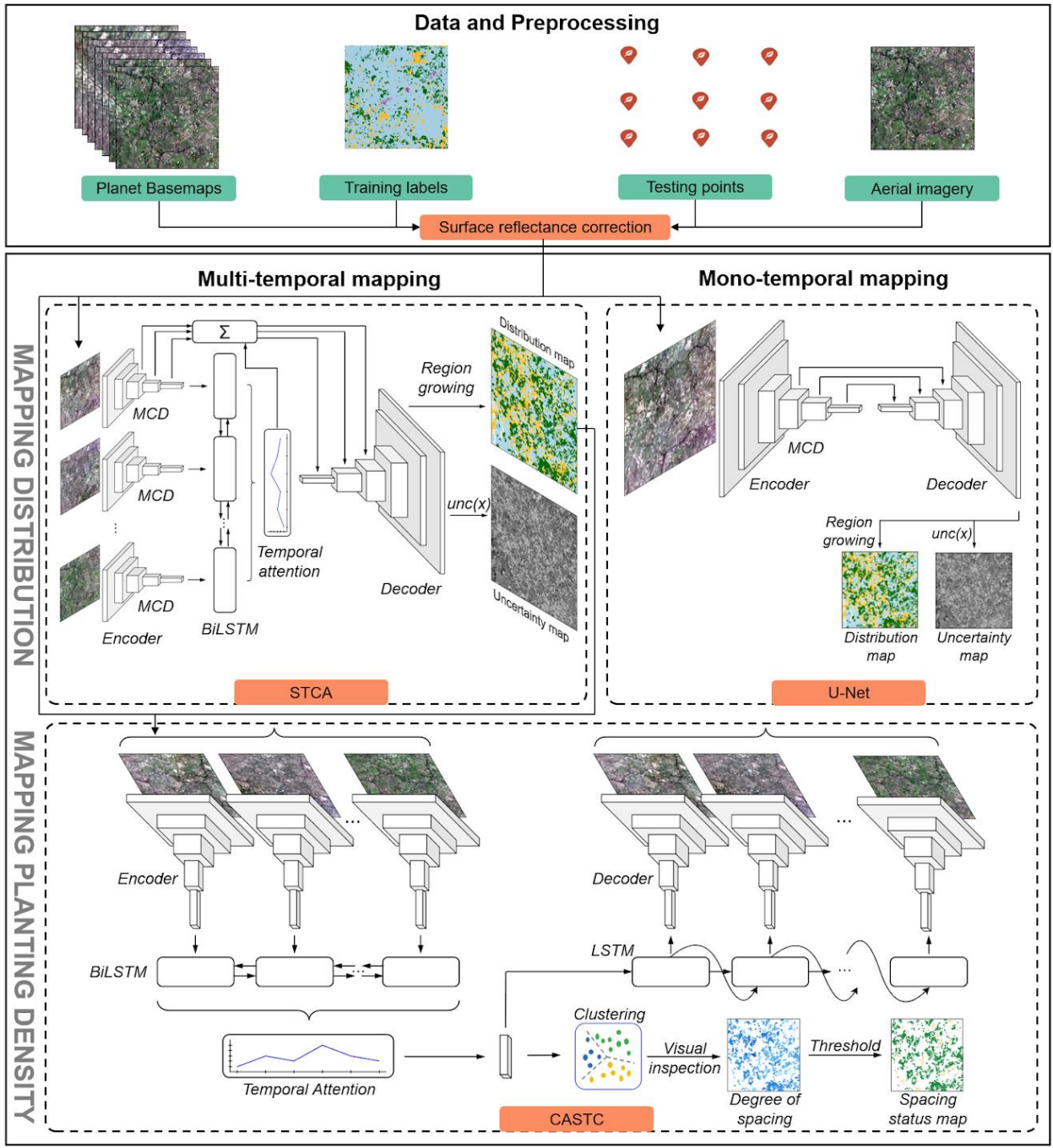

Fig. 3. Overview of tree crop mapping algorithms for cashew plantations along with the data and methods employed, and the maps generated. There are two branches for multi-temporal and mono-temporal imagery, respectively. Each deep learning module is enclosed by a dotted box. MCD indicates Monte Carlo dropout, and unc(x) is defined by Eq. (4).

This study proposes tree crop mapping algorithms for cashew plantations that address both the spatial distribution and planting density in two stages (Fig. 3). At the distribution mapping stage, the SpatioTemporal Classification with Attention (STCA) model was developed and U-Net (Ronneberger et al., 2015) was applied to map cashew plantation distributions for multi-temporal imagery from 2019 to 2021 and for mono-temporal imagery in 2015, respectively. At the planting density mapping stage, the Clustering Augmented Self-supervised Temporal Classification (CASTC) model was developed to map cashew planting density for multi-temporal imagery in 2021.

The training and testing steps using multi-temporal Planet Basemaps were conducted in a NVIDIA V100 GPU with 32 GB memory. Training and testing using mono-temporal aerial imagery were conducted in two NVIDIA V100 GPUs with 64 GB memory. For both sets of imagery, we used the same PyTorch deep learning framework.

**2.4.2 Mapping the distribution of cashew plantations**

2.4.2.1 Smoothing SR differences between tiles

Each Planet Basemap is a composite image made up of images from different microsatellites in the PlanetScope constellation, and therefore frequently has issues with inconsistent SR for the same ground object between tiles, even after the correction based on the monthly MODIS product (Houborg and McCabe 2018; Rao et al., 2021). In addition, various levels of clouds and shadows also lead to multiform SR for the same ground object in different regions, despite the superior image quality of Planet Basemaps relative to the SR products of Sentinel-2 and Landsat imagery in the tropics. Some studies have used other relatively mature SR products to correct Planet imagery to address these problems. For example, Sentinel-2 and Landsat SR products were used to stretch the histograms of Planet data (Jain et al., 2016; Rao et al., 2021). However, this method is unsuitable for our study region because Sentinel-2 and Landsat data have much more cloud coverage than Planet Basemaps in the tropics. Given that a small area has a high chance of being repeatedly captured by the same sensor under similar cloud and shadow conditions, we split our study region into many small areas. To smooth SR differences between tiles, we then performed a normalization on each of them using Eq. (1), where $P_{normalized}$ is the normalized pixel value, $P$ is the original pixel value, and $P_{min}$ and $P_{max}$ are the minimum and maximum pixel values respectively. In this study, we split our study region into 469 small areas according to the Planet Basemaps tile boundary for convenience. Note that we removed pixel values in the top and bottom two percentiles to avoid abnormal values. The same preprocessing was applied to aerial imagery.

$$P_{normalized} = \frac{P - P_{min}}{P_{max} - P_{min}} \quad (1)$$

2.4.2.2 Cashew plantation maps

We employed the STCA model to create cashew plantation maps for years from 2019 to 2021 using the Planet Basemaps. The STCA model can leverage the spatiotemporal information from time-series imagery to create a high-confidence cashew plantation map. Although several deep learning approaches that leverage spatiotemporal information for crop mapping (Ji et al., 2018; Jia et al., 2017; Mazzia et al., 2020) have shown encouraging results in isolated scenarios for studying a specific crop, these approaches used Convolutional Neural Networks (CNN) and Recurrent Neural Networks (RNN) in a straightforward manner to model spatial and temporal information without determining how various time steps contribute to the classification performance. Instead, we need a model that can automatically pay attention to the time steps that contribute the most to classification performance. The STCA model uses a U-Net-like module to automatically extract spatial features and a Bidirectional Long-Short Term Memory (BiLSTM) module (Graves and Schmidhuber 2005) to extract phenological changes (Ghosh et al., 2021). To better aggregate the information for each time step, we further added the attention mechanism to aggregate the hidden representations over the time series based on their contribution to the classification performance (Luong et al., 2015; Jia et al., 2019). The parameter set $\theta$ of STCA was trained to minimize the objective function of pixel-wise cross entropy on the limited number of manually annotated labeled patches:

$$\mathcal{L}(\theta \mid X, Y) = -\frac{1}{N \times 64 \times 64} \sum_{i=1}^{N} \sum_{(h,w)}^{(64,64)} \sum_{k=1}^{4} (Y_i)_{h,w}^k \log f(X_i; \theta)_{h,w}^k \qquad (2)$$

where, $N$ is batch size, $X_i$ and $Y_i$ denote the $i^{th}$ patch in the batch and its corresponding label of size $(h, w)$, and $f(X_i; \theta)$ is the predicted probabilities of each class $k$ on the $i^{th}$ patch. In this study, we used the kernel of 3 by 3 pixels for five convolution layers and the kernel of 2 by 2 pixels for max-pooling and transposed convolution layers in encoder and decoder.

For each year, imagery data for the seven months from November to May were fed into STCA. Before model training, each image tile was cut into image patches of 64 by 64 pixels. To create a more stable prediction result and quantify prediction uncertainty, Monte Carlo dropout was used in the testing phase. For standard deep learning models, dropout is only applied during training, which serves as a regularization to avoid overfitting. In our study, dropout was also applied in the testing phase. Specifically, we randomly sampled the neurons to be dropped out in each hidden layer, resulting in a slightly different model architecture, which can be viewed as an averaged ensemble of multiple different neural networks (Gal et al., 2016). During testing, the predicted probability value for each class was obtained by taking the average of ten runs, as demonstrated in Eq. (3):

$$f(X_i) = \frac{1}{10}\sum_{r=1}^{10} f(X_i; \theta_r) \qquad (3)$$

where $X_i$ denotes the $i^{th}$ patch and $f(X_i; \theta_r)$ is the predicted probability of each run r on the $i^{th}$ patch. Moreover, the standard deviation across the runs gives an estimate of the uncertainty unc in the prediction (Eq. (4)). In this study, the dropout rate is 0.3.

$$unc(X_i) = \sqrt{\frac{\sum_{r=1}^{10}(f(X_i; \theta_r) - f(X_i))^2}{10}} \qquad (4)$$

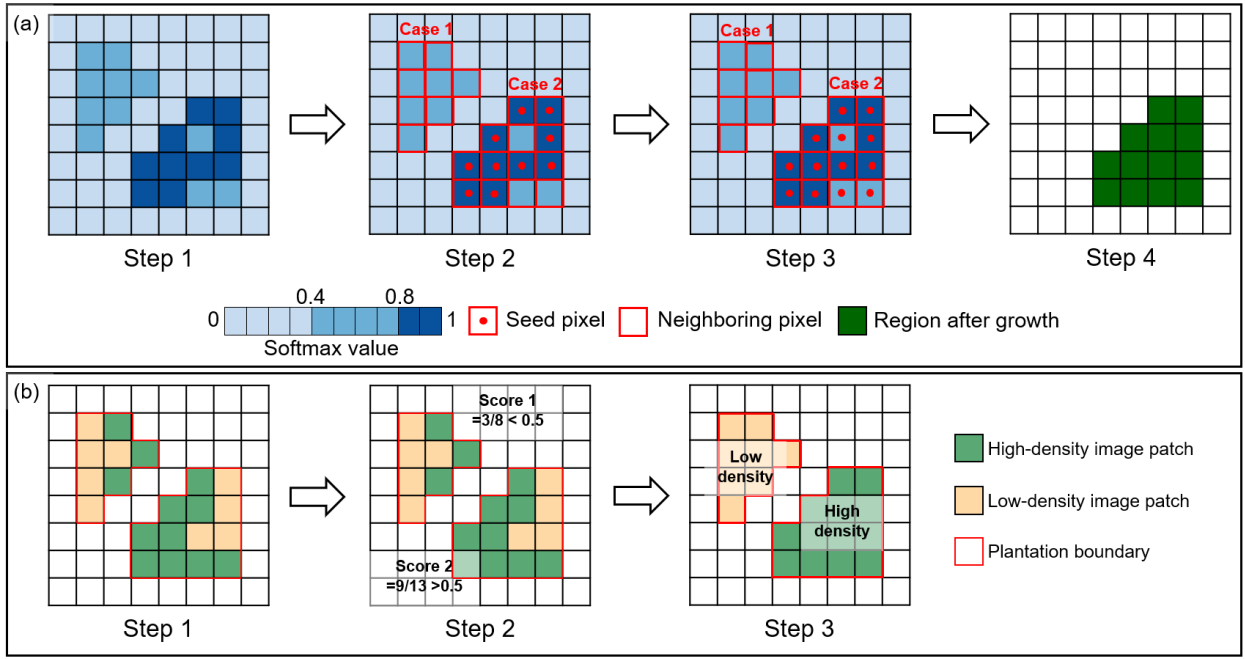

Fig. 4. (a) Region growing strategy and (b) the pathway to distinguish between high-density and low-density cashew plantations.

Next, each pixel was allocated a category using a region growing strategy. Traditionally, for deep learning multi-classification tasks, each pixel is assigned to the class with the highest Softmax value, and we refer to this strategy as "Argmax" prediction. However, with Argmax prediction, misclassification between spectrally similar classes and fragmentation of fields can easily occur. In our study, cashew plantations share similar spectral features with gallery forests and other tree crops, e.g., for some pixels which should be gallery forests or other tree crops, the highest Softmax value may be for cashew plantations. In this case, the Argmax prediction would wrongly classify these pixels as cashew plantation. In addition, the space between cashew trees in plantations may be identified as other land cover types, resulting in unrealistically fragmented fields. Therefore, the region growing strategy was chosen to process the pixel-wise Softmax output instead of directly taking Argmax prediction as a classification result. Specifically, step 1 in Fig. 4(a) shows the original Softmax output for one class, where darker blue colors represent larger Softmax values. As shown in step 2, pixels having a Softmax value greater than 0.8 were assigned as seed pixels, and those between 0.4 and 0.8 were assigned as neighboring pixels. In step 3, neighboring pixels with a seed pixel in their neighborhood were reassigned as seed pixels. Step 4 shows the region growing result generated by seed pixels. The neighboring pixels in case 1 would not be maintained in the end, which reduces misclassification, while the neighboring pixels in case 2 would be maintained to ensure the integrity of plots.

We employed the U-Net model to create the cashew plantation map for 2015 using aerial imagery. Each image tile was cut into image patches of 256 by 256 pixels before model training. The U-Net model has been prevalent in the crop classification domain (M and Jayagopal 2021; Wei et al., 2019; Zou et al., 2021) because of its advantages for segmentation tasks, including the combined use of global location and context, fewer training samples, and good performance (Alom et al., 2018; Ronneberger et al., 2015). The same Monte Carlo dropout method was applied to create an average over ten predictions and express the prediction uncertainty. In this study, we used five convolution layers with the kernel of 3 by 3 pixels. Maxpool and transposed convolution layers were employed with the kernel of 2 by 2 pixels. The dropout rate is set to 0.3. Then, the region growing strategy was applied to produce the final cashew plantation map for 2015.

2.4.2.3 Classification post-processing

Based on the fact that perennial cashew plantations are not cut down once planted in our study region, the cashew plantation area identified in 2015 was used as a mask in the 2019, 2020, and 2021 classification maps to avoid omission errors. Similar masks were generated for 2019 and 2020 to identify cashew plantations over the years that followed. Given the fair classification performance for the built-up land class due to limited training labels, we used the land cover maps published in 2016 (ESA, 2017) and 2020 (ESA, 2021) by ESA to create masks for built-up land to update the classification result.

**2.4.3 Mapping cashew planting density in plantations (tree-spacing practices)**

Field-collecting tree planting density data on a large scale is a difficult task requiring a substantial commitment of time and resources, which renders supervised learning unrealistic. Therefore, we need a learning strategy to categorize plantations into high- or low-planting density without the need of labels. Clustering is one such widely used unsupervised learning strategy. However, directly clustering the spectral bands of mono-temporal remote sensing imagery can lead to suboptimal results for several reasons. First, the clustering using spectral values at the pixel level can be noisy due to the spatial noise of the sensors. Second, clustering using all spectral values of all the pixels in an image patch can lead to very high dimensions and thus lead to issues like correlated attributes and inaccurate calculated distances. Third, multi-temporal remote sensing imagery includes more information to depict cashew tree crops than mono-temporal imagery. Thus, to address these challenges, we extracted abstract features – embedding – from the satellite imagery time series using a spatiotemporal autoencoder and applied the clustering method at the embedding level. Furthermore, the traditional K-means clustering method (MacQueen et al., 1967) cannot perform well with complicated imagery datasets (Xie et al., 2016), and it likewise fared poorly in our task. Therefore, we adopted a deep embedded clustering method (Xie et al., 2016) as our clustering method, which optimized the K-means result and thus improved our clustering performance (Ghosh et al., 2022).

In this study, the Clustering Augmented Self-supervised Temporal Classification (CASTC) model was leveraged to distinguish two kinds of cashew plantation tree-spacing practices (high-density versus low-density cashew plantations) using the temporal Planet Basemaps in 2021. Specifically, the encoder and decoder parts in the autoencoder structure have a similar architecture to STCA without the skip connections. Removing skip connections can help the model to extract quality encoded spatiotemporal vectors – embedding – that fully capture representative features without the assistance from the skip connections. The embedding output by the encoder is fed into an LSTM-based sequence decoder that generates a sequence of vectors, and a convolutional decoder then reconstructs back the input time-series satellite imagery based on the vector sequence. Model training consists of two phases: the first phase involves model initialization, and the second phase is model refinement with optimized clustering objectives. In the first phase, the autoencoder generated the embeddings, which were then subjected to the K-means clustering algorithm to generate initial cluster centroids. In the second phase, the cluster centroids obtained from the first phase were refined by matching soft assignment to target distribution by Kullback-Leibler (KL) divergence (Joyce, 2011) minimization. The soft assignment measures the similarity between an embedding and a cluster centroid with the t-distribution (Van der Maaten et al., 2008) to generate the probability of assigning the embedding to the cluster:

$$q_{ij} = \frac{\left(1 + |h(X_i; \theta_h) - M_j|^2/\alpha\right)^{-\frac{\alpha+1}{2}}}{\sum_{j'=1}^{K}\left(1 + |h(X_i; \theta_h) - M_{j'}|^2/\alpha\right)^{-\frac{\alpha+1}{2}}} \tag{5}$$

where $q_{ij}$ is the probability of assigning the $i^{th}$ embedding to the $j^{th}$ cluster, $h(X_i; \theta_h)$ is the $i^{th}$ embedding, $M_j$ is the $j^{th}$ cluster centroid, α is the degree of freedom, and K is the number of cluster centroids. To promote model learning from high-confidence embeddings, the target distribution was computed by skewing the soft assignment to drive the embedding $i$ closer to the cluster $j$ with highest $q_{ij}$:

$$p_{ij} = \frac{q_{ij}^2/\sum_i q_{ij}}{\sum_{j'=1}^{K}\left(q_{ij'}^2/\sum_i q_{ij'}\right)} \tag{6}$$

Then, we matched the soft assignment to the target distribution by minimizing the KL divergence to refine cluster centroids and encoder:

$$\text{KL} = \frac{1}{N_t}\sum_{i=1}^{N_t}\sum_{j=1}^{K} p_{ij} \log\left(\frac{p_{ij}}{q_{ij}}\right) \tag{7}$$

where $N_t$ is the number of image patches in the training set.

Before model training, each image tile was cut into image patches of 32 by 32 pixels. After training the model, we had ten clusters, and the model was used to assign a cluster for each image patch in the training set. For the image patches for each cluster, we then visually inspected the corresponding high-resolution Airbus satellite imagery and assigned the image patches as high density or low density (step 1 in fig. 4(b)). We applied this model to the entire study region and kept only the cashew plantation region using a cashew plantation distribution mask. Each separate cashew plantation was given a density score from 0 to 1 (step 2 in Fig. 4(b)) to indicate the ratio of high-density image patches according to Eq. (4). Then, a density score threshold of 0.5 was applied to distinguish between high-density and low-density plantations (step 3 in Fig. 4(b)).

$$density\ score = \frac{N_{high-density-patches}}{N_{all-patches}} \qquad (8)$$

**2.4.4 Validation of cashew plantation classification maps**

Two accuracy assessment methods were performed on the cashew plantation distribution in two dimensions: space and time. In the spatial dimension, we verified the accuracy for each year using 1,400 ground truth samples. A confusion matrix, overall accuracy (OA), user's accuracy (UA), and producer's accuracy (PA) were used to evaluate the accuracy. In the time dimension, we verified the consistency of the classification results from 2015 to 2021 using 196 cashew plantation samples located in the same place across years. Note that built-up areas were ignored in our accuracy assessment, as this information was derived from existing products and was not the focus of this study. To assess the accuracy of cashew plantation tree-spacing practices, 348 points with known planting density were used.

**2.4.5 Benchmarking with other classification approaches**

The performance of STCA was compared with three deep learning methods: 3D-CNN (Ji et al., 2018), U-Net with ConvLSTM (Ghosh et al., 2021; Shi et al., 2015), and CALD (Jia et al., 2019), all of which were designed to learn spatiotemporal information and have been shown to perform well in crop classification tasks. We did not include traditional machine learning methods for comparison with STCA because they do not take spatial information into account in their classification, and the CALD approach has been shown to perform much better on cropland mapping than RF and SVM (Jia et al., 2019). Compared to the traditional CNN network, 3D-CNN exploits additional temporal information by conducting convolution in the time dimension to learn temporal information. U-Net with ConvLSTM utilizes the U-Net architecture but substitutes the convolution layers with ConvLSTM layers, which can capture spatial and temporal information with CNN and LSTM. CALD leverages a context-aware LSTM to capture temporal information and further learn spatial information from neighboring pixels. We compared the classification performance of these four approaches within the training region. UA and PA were used to assess classification performance for each class.

We also compared the performance of CASTC with two other standard self-supervised classification methods, i.e., Autoencoder with K-means (Ghosh et al., 2022) and Colorization with K-means (Vincenzi et al., 2021). Autoencoder is a standard spatiotemporal STCA architecture without the skip connections. Colorization is a self-supervised learning technique with two independent branches taking in the NIR and RGB channels. Both of the branches are trained by the autoencoder separately, and we averaged their respective embeddings from the final layer as final embeddings. K-means is performed on the embeddings from the final layer of the encoder for both methods. To compare the clustering performance of the three approaches at the embedding level, we measured the inter- and intra-cluster differences using two metrics, the Separability Index ($SI$) and Coefficient of Variation ($CV$). $SI$ (Somers and Asner et al., 2013) was used to measure the inter-class difference (Eq. (9)):

$$SI_{i,j} = \frac{|\mu_i - \mu_j|}{\sigma_i + \sigma_j} \tag{9}$$

where i and j represent different embedding clusters; $\mu_i$ and $\mu_j$ refer to the mean value of cluster i and cluster j, respectively; $\sigma_i$ and $\sigma_j$ represent the standard deviation of cluster i and cluster j, respectively. The numerator can reflect the disparity between different clusters, while the denominator can indicate the degree of concentration within clusters (Hu et al., 2019; Yin et al., 2020). A larger $SI$ indicates greater dissimilarity between the embeddings in the two clusters. $CV$ was used to measure intra-cluster differences, and it is unitless (Eq. (10)):

$$CV_i = \frac{\sigma_i}{\mu_i} \tag{10}$$

where i represents a cluster, and $\sigma_i$ and $\mu_i$ refer to the mean value and standard deviation of a cluster, respectively. A smaller $CV$ indicates a greater concentration of embeddings in the cluster. Then, we compared the distribution of $SI$ and $CV$ for the three approaches.

## 3 Results

### 3.1 Results of benchmark experiments

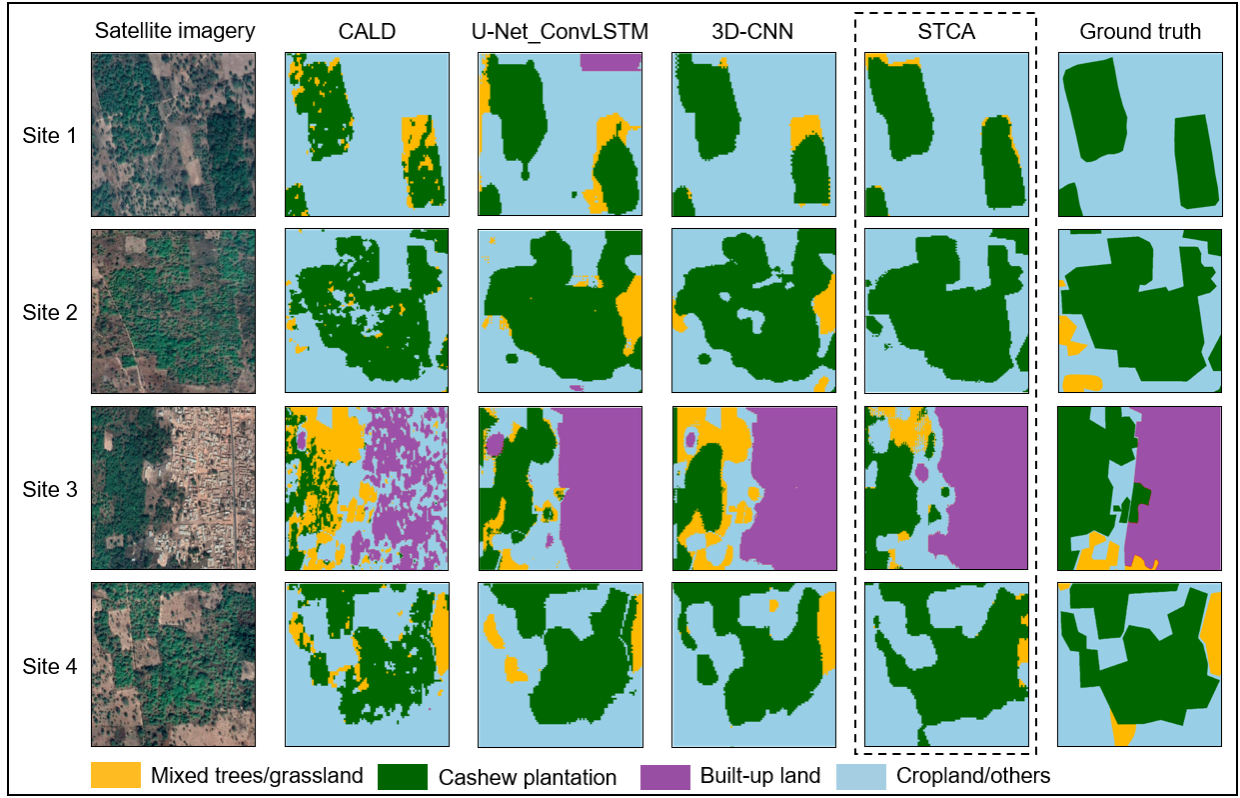

Fig. 5. Classification results for STCA and baselines along with ground truth labels and high-resolution satellite imagery from Google Earth.

The accuracy assessment results in Table 2 compare the classification performance between STCA and the other three benchmark approaches. The classification accuracy of the STCA method for cashew plantations is superior to the other three methods. CALD exhibited the poorest performance because it didn't capture enough spatial information, which results in more errors at individual pixels. 3D-CNN and ConvLSTM demonstrated slightly better classification abilities than CALD, although both methods are unable to identify the time steps that contribute most. Fig. 5 provides a visual comparison of the classification results. The CALD method classification results are too fragmented to maintain continuity of different land covers. The ConvLSTM and 3D-CNN generate more misclassification between cashew plantations, mixed trees/grassland, and cropland/others than STCA.

Table 2. Accuracy assessment for STCA and benchmark methods.

| PA/UA \ Methods | CALD | U-Net+ConvLSTM | 3D-CNN | STCA |
|---|---|---|---|---|
| Mixed trees/grassland (PA) | 81.6% | 82.4% | 83.5% | 88.1% |
| Mixed trees/grassland (UA) | 82.3% | 83.2% | 81.9% | 84.4% |
| Cashew plantation (PA) | 75.1% | 77.2% | 78.4% | 85.7% |
| Cashew plantation (UA) | 76.3% | 78.5% | 77.2% | 83.0% |
| Built-up Land (PA) | 63.4% | 65.5% | 58.3% | 70.4% |
| Built-up Land (UA) | 50.7% | 55.9% | 60.7% | 65.5% |
| cropland/others (PA) | 68.3% | 70.4% | 74.8% | 80.2% |
| cropland/others (UA) | 72.3% | 73.6% | 69.6% | 84.7% |

Fig. 6(a) illustrates the classification comparison between CASTC and the two other benchmark self-supervising approaches. In terms of $SI$, although CASTC has a slightly lower median value than Colorization with K-means, the first quantile and the maximum are noticeably higher, which indicates much greater divergence of some cluster pairs. Autoencoder with K-means has the poorest clustering performance. In the comparison of $CV$, the maximum and median of CASTC are the lowest among the three methods, although the minimum is slightly higher than Colorization with K-means. Colorization with K-means has the greatest median and maximum values of the three approaches, indicating that more than half of the clusters it formed were highly dispersed. As shown in Fig. 6(b), the image patches from two clusters formed by CASTC show obvious differences in cashew tree density. High-density clusters stand out in sharp contrast to low-density clusters. However, for Autoencoder and Colorization with K-means, high-density and low-density cashew plantations are mixed in the same cluster, indicating that the two methods perform poorly in this task.

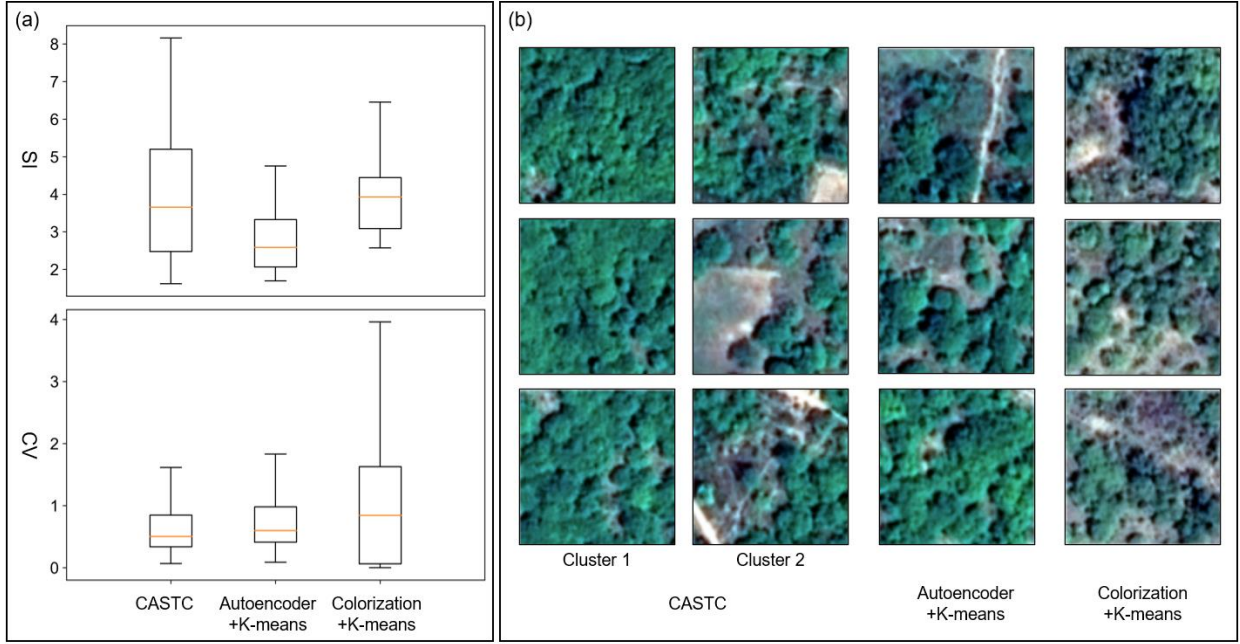

Fig. 6. Comparison of clusters generated by CASTC and benchmark methods by (a) SI and CV and (b) visual check.

## 3.2 Accuracy assessment for cashew plantation distribution and plantation density

The individual OA values of the cashew plantation maps are 0.89 ± 0.0186, 0.87 ± 0.0187, 0.85 ± 0.0204, and 0.90 ± 0.0164 for 2015, 2019, 2020, and 2021, respectively, (Table S1-4) with 1,400 validation points. The cropland/others class was most frequently misclassified with cashew plantations. A small amount of Mixed trees/grassland labels were also found in points predicted to be cropland/others. The greater UA of cashew plantations for 2019, 2020, and 2021 than in 2015 was expected, given that no temporal information could be used for the mono-temporal imagery. In terms of classification consistency, 83.7% of the samples were consistently classified as cashew plantation across the years, which demonstrated the stability of the classification along the time dimension (Table S5). The OA for cashew plantation density mapping was 0.76 ± 0.0492 (Table S6). Low-density cashew plantations were better categorized than high-density plantations, possibly because the number of high-density plantation samples, 128, was smaller than that of low-density samples, 220.

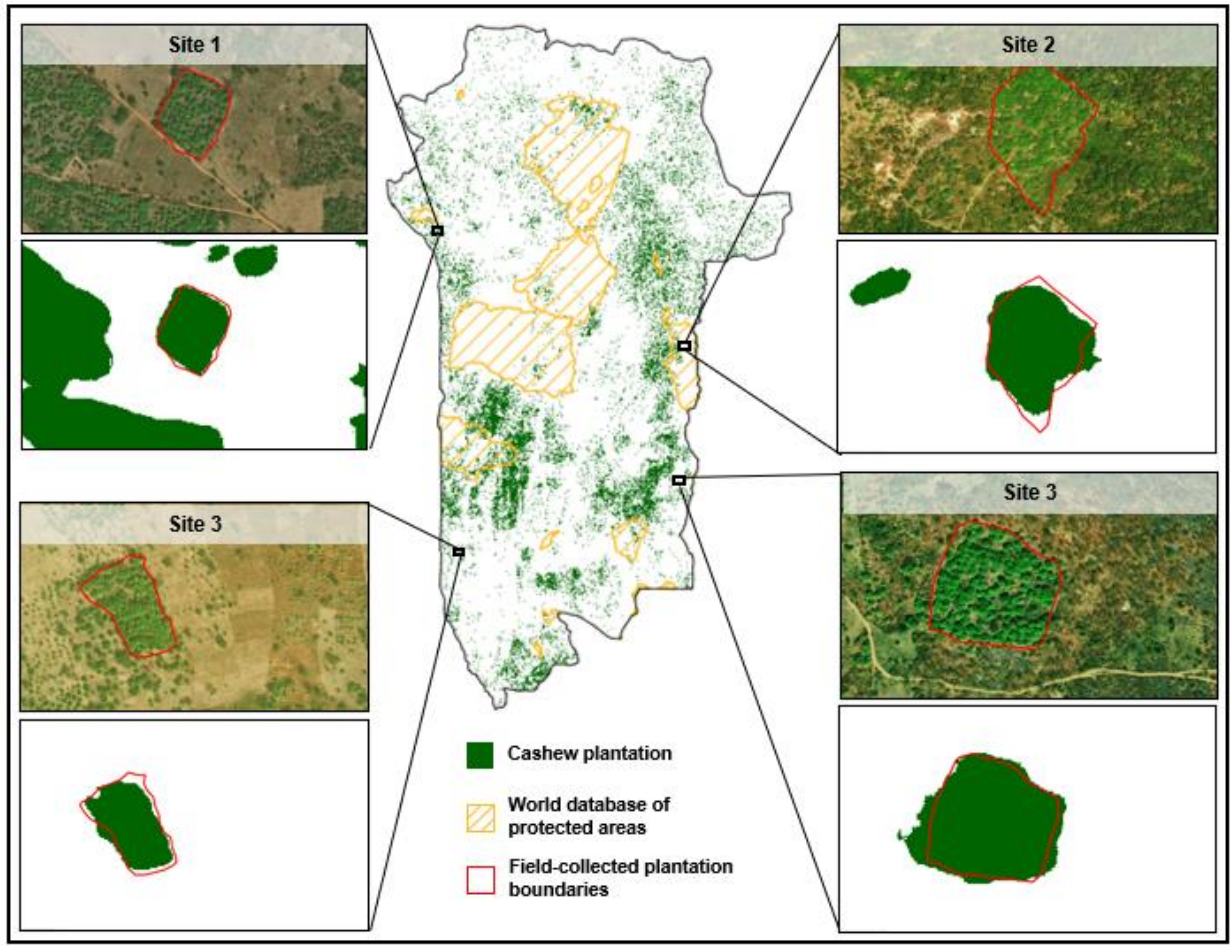

Fig. 7. The distribution of cashew plantations in 2021. Three sample sites are shown with classification maps and satellite imagery from Google Earth.

**3.3 Spatial distribution of cashew plantations in 2021**

Fig. 7 shows the distribution of cashew plantations in 2021, and Table 3 summarizes the cashew plantation area and share of land planted with cashew trees in 2021 for each commune. The area of cashew plantations in 2021 is 519 ± 20 thousand hectares (kha). The Bantè commune has the greatest share of land dedicated to cashew plantations, with nearly 300 ha of plantations per kha land, while two other communes (Savalou and Djidja) have nearly 200 ha of cashew plantations per kha land. Djougou, N'Dali, and Bassila communes have less than 100 ha cashew plantations per kha land. The Tchaourou has the greatest total cashew plantation area with 102 ± 7 kha, and account for nearly one fifth of total cashew plantation area in Benin. Additionally, we found that cashew plantations have encroached on 45 ± 6 kha of protected areas, using boundaries in Benin within the World Database of Protected Areas (WDPA). 55% more protected area has been encroached upon by cashew plantations in 2021 than in 2015 (29 ± 7 kha). The online version of the cashew plantation map is available on TechnoServe Lab (https://cajuboard.tnslabs.org).

Table 3. Total cashew area and share of land planted with cashew trees for communes.

| Communes / Metrics | Cashew area (kha) | Share of land planted with cashew trees (ha/kha) |
|---|---|---|
| N'Dali | 35±6 | 90±15 |
| Parakou | 7±1 | 149±21 |
| Tchaourou | 102±7 | 140±10 |
| Bantè | 79±4 | 299±15 |
| Dassa-Zoumè | 20±5 | 114±28 |
| Glazoué | 23±5 | 142±28 |
| Ouèssè | 43±6 | 174±19 |
| Savalou | 46±7 | 197±26 |
| Savè | 51±6 | 128±26 |
| Bassila | 40±5 | 40±10 |
| Djougou | 29±4 | 73±10 |
| Djidja | 35±9 | 199±43 |

**3.4 Cashew plantation expansion from 2015 to 2021**

The classification results for the four years indicate that the area of cashew plantations almost doubled from 268 ± 15 kha to 519 ± 20 kha between 2015 to 2021. This confirms public statistics from the Benin Ministry of Agriculture, Livestock and Fisheries that the area under cultivation increased by 71% from 286 kha in 2016 to 488 kha in 2020 (MAEP-Benin, 2020). During the last seven years, there were clear signs of growth in west and east Collines, west Donga, and south Zou, as shown in Fig. S2. Fig. 8(a) quantitatively shows the composition of the original land cover type from which the new cashew plantations were established. The departmental statistics show that the cashew plantation area increased in all four departments over the past seven years with the highest growth in Collines and the lowest growth in Zou. From 2015 to 2021, the cropland/others class accounted for more cashew expansion than the mixed trees/grassland class for all four departments. Only in two instances – in Borgou and Zou between 2015 and 2019 – did conversion to cashew plantations from mixed trees/grassland exceed 50%. The highest conversion from mixed trees/grassland to cashew plantations occurred in Borgou from 2015 to 2019 with 58.3% cashew expansion, while the lowest occurred in Zou from 2020 to 2021, with 23.4% cashew expansion.

The statistics of the same metrics are provided for each commune in Fig. 8(b). All 12 communes had continuous cashew expansion over the last seven years. The highest cashew plantation area growth occurred in Savè, while the lowest growth occurred in Parakou, likely because the area of Parakou is smallest among the 12 communes and space for establishing new cashew plantations is very limited. The relative conversion of mixed trees/grassland and cropland/others to cashew plantations varies greatly from region to region. Between 2015 to 2019, more mixed trees/grassland than cropland/others were converted to cashew plantations in Bantè, Ouèssè, Savalou, Djidja. Two such individual changes also occurred each between 2019 to 2020 (Bassilla and Dassa-Zoumé) and between 2020 to 2021 (Bassilla and Parakou). For other periods and communes, the cropland/others class accounted for more cashew expansion than mixed trees/grassland.

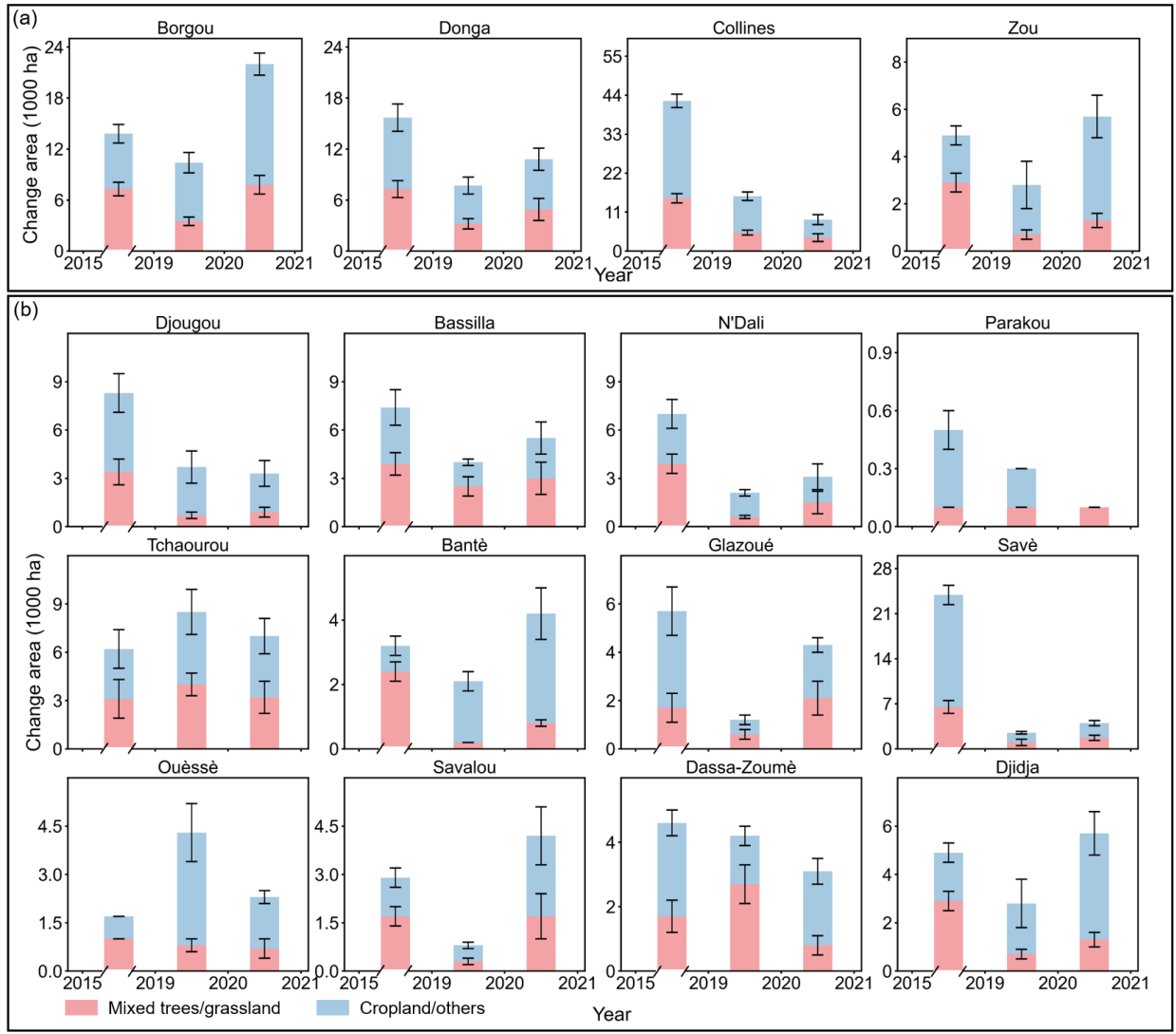

Fig. 8. The source of cashew plantation area growth at (a) the department level and (b) the commune level. Note that the change between 2015 and 2019 represents a larger time increment on the x-axis than between individual years.

**3.5 Planting density map of cashew plantations for 2021**

The map of cashew plantation planting density scores is shown in Fig. 9(a). A threshold planting density score of 0.5 was selected to distinguish high-density and low-density cashew plantations (Fig. 9(b)), and details are shown for three sample sites in Fig. 9(c). As shown in Fig. 9(d), the proportion of high-density cashew plantations relative to the total reveals that 5 communes exceed 50%, namely N'Dali, Parakou, Bantè, Ouèssè, and Savalou. In the Tchaourou commune, nearly half of the area planted with cashew trees already consists of high-density cashew plantations. These six communes are all located in the Borgou and Collines departments, which means these two departments, especially Borgou, have a relatively mature cashew cultivation industry. On the other hand, although the three communes of Savè, Bassilla, and Djougou have relatively large cashew plantation areas, the tree density is low, which presents an opportunity to increase cashew planting density (and therefore cashew production) on existing plantations.

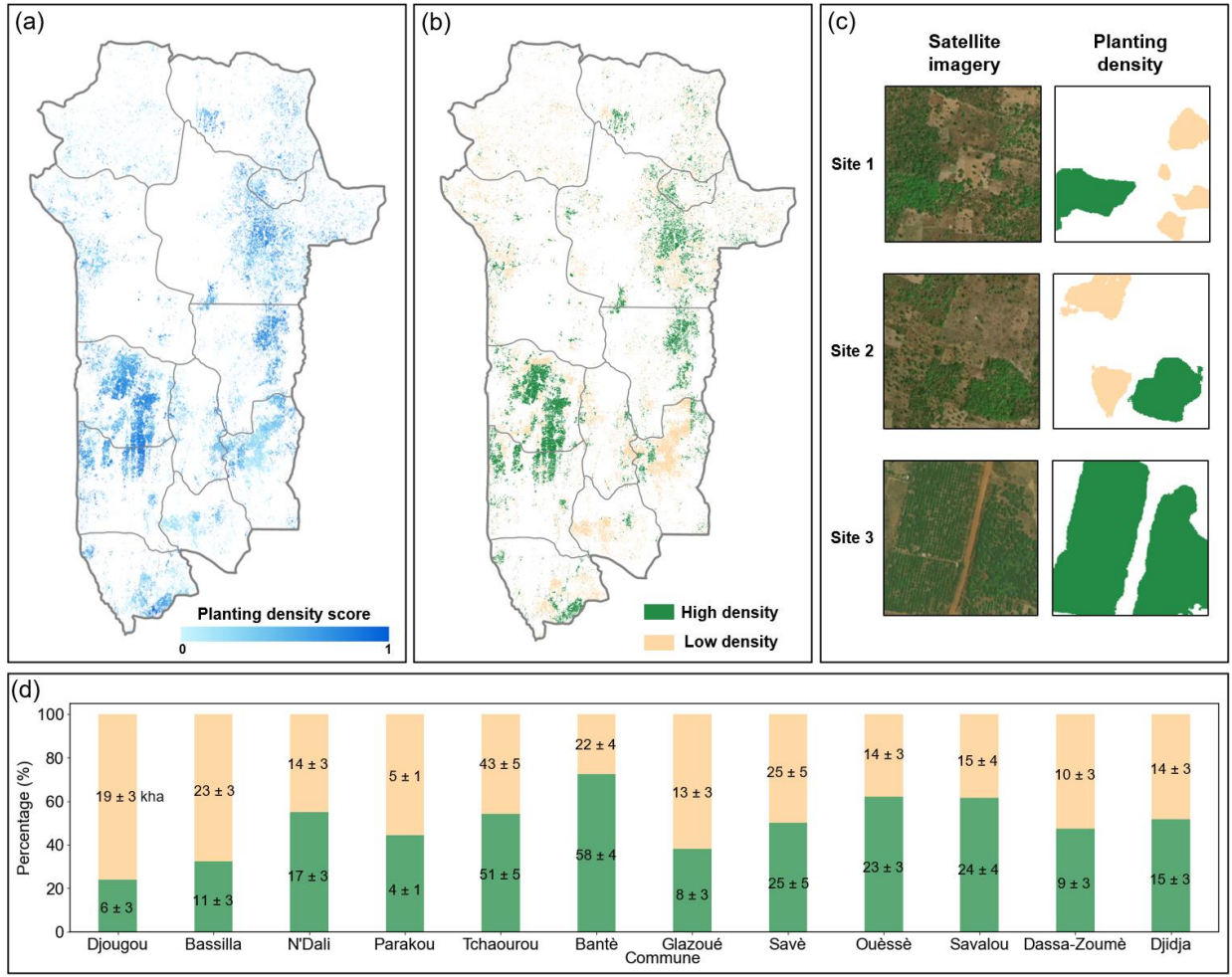

Fig. 9. (a) A map of cashew plantation planting density scores in 2021. (b) A map of high- and low-density cashew plantations in 2021 with a threshold of 0.5. (c) Three sample regions with high- and low-density cashew plantations. The high-resolution satellite imagery is from Google Earth. (d) The percentage and area of high- and low-density cashew plantations for all communes.

## 4 Discussion

### 4.1 The added value of Planet Basemaps in large-scale smallholder crop mapping

Although some previous studies have focused on smallholder crop mapping, the limited spatial and temporal resolution of the available remote sensing data was not adequate for creating field-level maps. Prior to the Sentinel mission launch, Landsat and MODIS data were used to map smallholder farms (Jain et al., 2013; Schneibel et al., 2017). Later, Sentinel-1/2 were widely used for crop functional mapping of all kinds, especially in Africa (Jin et al., 2019; Lambert et al., 2018; Masiza et al., 2020). However, the highly fragmented fields and frequent cloud coverage in some regions can cause satellite imagery with medium spatial/temporal resolution to lose its efficiency. Sentinel-2 imagery cannot adequately depict smallholder cashew plantation field boundaries, given its insufficient spatial and temporal resolution and the degradation of image quality by clouds and shadows (Fig. S3(a)). Recently, researchers have realized the advantages of the high spatial and temporal resolution in Planet's microsatellite constellation for crop mapping. Some promising smallholder crop mapping results using Planet Daily Scenes SR data (3 m) have been published, although they consist solely of small-scale ($< 1000 \text{ km}^2$) applications (Rafif et al., 2021; Rao et al., 2021). However, such daily images are still affected by clouds and shadows to varying degrees (Fig. S3(b)). In comparison, Planet Basemaps generally contain less noise (Fig. S3(c)). This is because the monthly Planet Basemaps go through more post-processing steps than Planet Daily Scenes. First, the cloud cover and image sharpness are used as quality metrics to determine the best imagery in each month, and the imagery that ranks highest in the weighting of these metrics is selected. Then, the selected imagery is normalized to a monthly MODIS SR target to minimize variability between scenes and reduce atmospheric effects. Finally, all images processed for Planet Basemaps are manually inspected for quality (Planet, 2022b). To summarize, Planet Basemaps enable field-specific, and even sub-field, crop monitoring.

We compared the classification performance for Sentinel-2 Level-2A, Planet Daily Scenes, and Planet Basemaps in the training region (Fig. 10). The classification result generated by Sentinel-2 was only able to identify a small subset of cashew plantations. Planet Daily Scenes were able to capture more cashew plantations than Sentinel-2, although some pixels were misclassified as cashew plantations. By comparison, Planet Basemaps classified cashew plantations more accurately. The classification accuracy assessment using F1 scores also showed that Planet Basemaps produced superior classification results compared to the other two methods (Table 4).

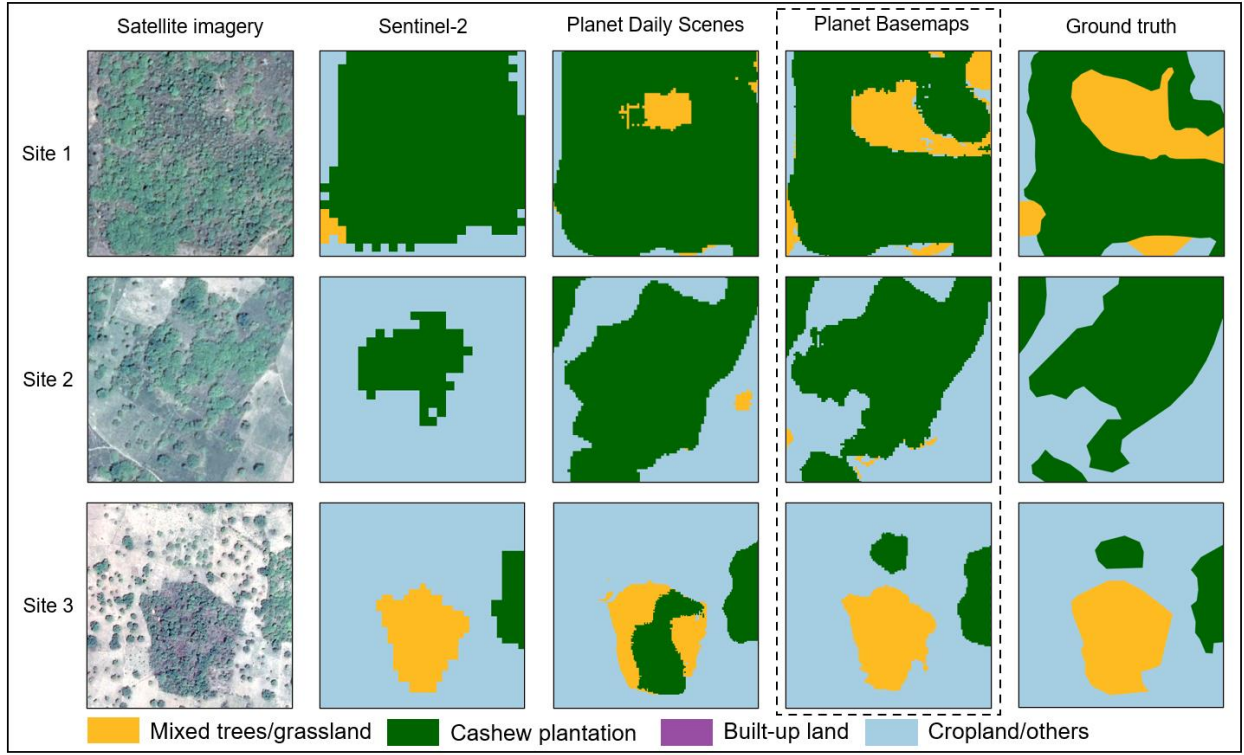

Fig 10. A detailed comparison of classification using Sentinel-2 imagery, Planet Daily Scenes, and Planet Basemaps, shown along with Airbus satellite imagery and ground truth labels.

Table 4. Accuracy assessment for Sentinel-2 Level-2A, Planet Daily Scenes, and Planet Basemaps using F1 score.

| Products  Class | Sentinel-2 Level-2A | Planet Daily Scenes | Planet Basemaps |
|---|---|---|---|
| Mixed trees/grassland | 0.7471 | 0.8535 | 0.9103 |
| Cashew plantation | 0.6723 | 0.7941 | 0.841 |
| cropland/others | 0.6837 | 0.7529 | 0.783 |

## 4.2 Uncertainty analysis of classification results

Because the study region is located in the tropics, clouds and shadow weaken the observational capability of optical sensors. At the same time, the Planet microsatellite constellation consists of many satellites, which inevitably causes differences in sensor characteristics, thus affecting the ability to obtain consistent SR for a large region to some extent. Additionally, other factors such as satellite product versions, atmospheric and directional corrections, and BRDF effects can also cause classification uncertainty (Zeng et al., 2022). All of these factors not only have impacts on the direct monitoring ability for cashew plantations, but also introduce inter-class similarity and intra-class differences, resulting in poor classification performance. In order to explore the impacts of these factors on the classification results, we employed an uncertainty mask generated by the Monte Carlo dropout technique (Eq. (4)) to filter out the pixels impaired by these factors. Fig. 11(a) shows the real surface conditions with satellite imagery. The pixel-wise classification uncertainty mask from ten runs with Monte Carlo dropout (Fig. 11(c)) was applied to the averaged classification results of the ten runs (Fig. 11(b)) with a threshold of 0.06 to generate a cashew plantation map without high-uncertainty pixels (Fig. 11(d)). The threshold can be adjusted case by case.

We took the classification result in 2021 as an example and divided the study region into three parts based on the degree of cashew plantation area decline (Fig. 11(e)) after applying the uncertainty mask. Communes are relatively spatially concentrated for each class. There are four communes in the southeast of the study region that experienced less decline than other communes (i.e., Djidja, Dassa-Zoumè, Savè, and Glazoué) among which Savè declined the least. On the other hand, the three northwestern communes of Djougou, Bassila, and Bantè declined more than other communes, among which Bassila declined the most.

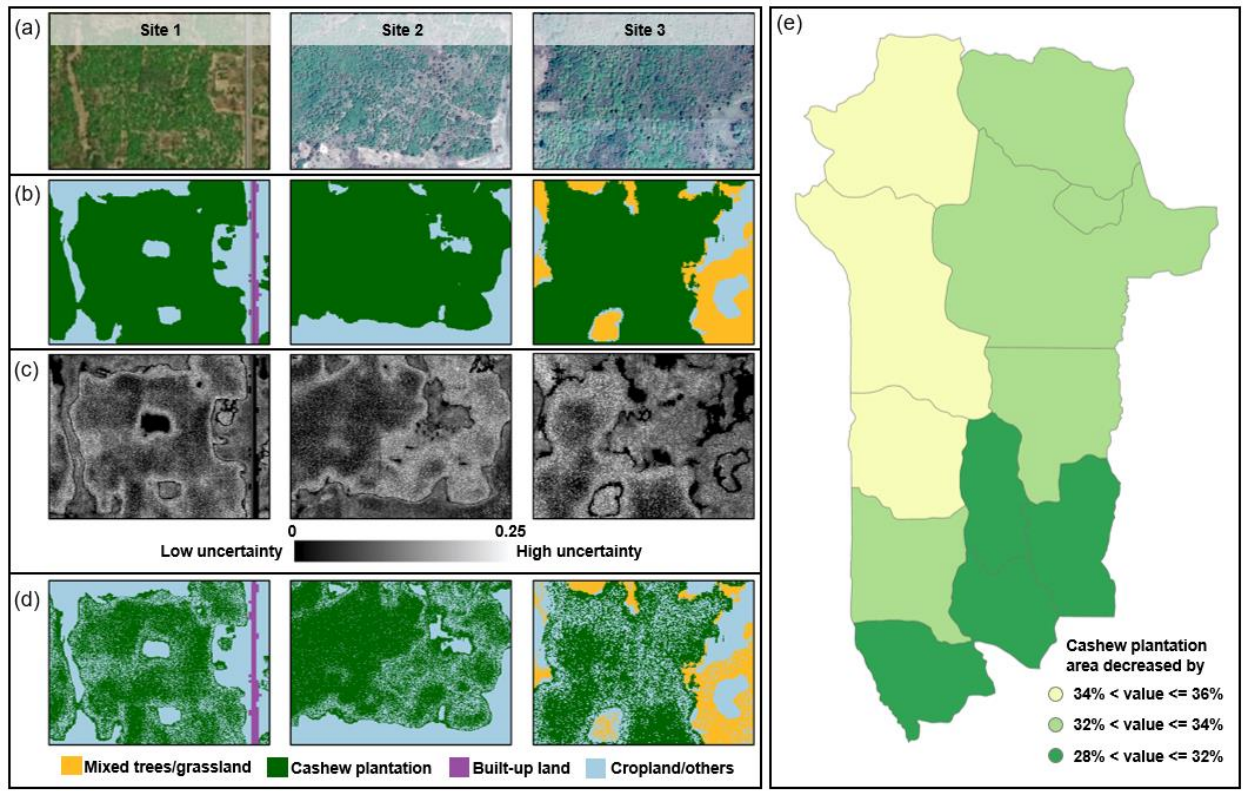

Fig. 11. The process to generate a cashew plantation map without high-uncertainty pixels. (a) Airbus/Maxar satellite imagery with 0.5-m resolution. (b) Classification map. (c) Uncertainty map. (d) The classification map masked by the uncertainty map with a threshold equal to 0.06. (e) Cashew area changes compared with the original classification map for each community after applying the uncertainty mask for 2021.

**4.3 Policy guidance for the cashew industry in Benin**

The findings of this study help document the progress of a major governmental initiative known as the Strategic Plan for the Development of the Agricultural Sector (PSDSA). A key goal under this plan was to double cashew production from 112,000 metric tons to 200,000 metric tons from 2016 to 2021 (MAEP-Benin, 2017). The results produced here provide additional inputs to inform the PSDSA for 2022-2025 (PNIASAN-Benin, 2022).

First, we tracked the increase in cashew areas under cultivation from 2015 to 2021. Our modeling shows that Benin increased cashew plantations from 268 ± 15 kha to 519 ± 20 kha between 2015 to 2021, which is very close to the government estimate of 286 kha under cultivation as of 2016 and 488 kha as of 2020 (MAEP-Benin, 2020). In 2021, Tchaourou had the largest cashew area (102 ± 7 kha) in 12 communes, and Parakou had the lowest (7 ± 1 kha). 60% of new cashew plantations came from lands that were previously under crop production or left fallow. According to the boundary of WDPA, the area of cashew plantations within protected areas grew from 29 ± 7 kha to 45 ± 6 kha between 2015 and 2021, an increase of 55%.

We further explored the overall planting density of cashew plantations based on the 2021 classification map. To increase cashew nut yield, it is important to implement good agricultural practices and to increase tree-planting density in areas we have classified as low-density. Our result shows that for the communes N'Dali, Parakou, Bantè, Ouèssè, and Savalou, more than 50% of the cashew plantation area is high-density. However, for the communes Djougou, Bassila, Glazoué, Savè and Dassa-Zoumè, most cashew plantations are low-density. In our study region, over 90% of new cashew plantations established between 2015 and 2021 were low-density. Increased application of inputs (seedlings) in low-density cashew plantations coupled with targeted training efforts in all cashew growing areas that have not yet received training could have a significant impact on cashew nut yields.

### 4.4 Limitations and future work

The combination of tree crop classification algorithms for cashew and high-resolution imagery demonstrated the power of accurately mapping the distribution and planting density of cashew plantations, upon which we monitored the dynamics of cashew areas under cultivation from 2015 to 2021. The proposed classification algorithms will allow rapid mapping of cashew plantations going forward. However, some limitations remain to be optimized in the future. First, because of year-round clouds and shadows in the tropics and differences in sensor characteristics of the satellite constellation, intra-class spectral inconsistency is still an issue for some regions, even after Planet Basemaps have been normalized to a monthly MODIS SR product. This is common in the remote sensing field, and solutions are limited. A potential method is using domain adaptation, including invariant feature selection, representation matching, adaptation of classifiers, and selective sampling (Elshamli et al., 2017; Martini et al., 2021; Tuia et al., 2016). Recently, the new generation of PlanetScope instruments with 4 newly added bands (coastal blue, yellow, a second green, and red edge spectral bands) has launched and started publishing imagery (Planet, 2022c). The new analytic product is calibrated to Sentinel-2 and has improved alignment, enabling accurate time-series analysis and machine learning models (Planet, 2022d). In future classification map updates, we will consider these new methods and inputs. Second, although the 2.4-m Planet Basemaps were leveraged to distinguish high- and low-density cashew plantations using the threshold value of 100 trees/ha, remote sensing imagery with higher spatial resolution can help differentiate additional levels of planting density (e.g., "very-high-density" plantations above 180 trees/ha) as a result of finer observational ability. In addition, more information about GAP adoption in cashew plantations can be captured, such as whether trees are pruned (i.e., tree crowns are not touching each other), which is important for improving cashew tree yields. In the future, more agricultural practices will be mapped.

With further work and close coordination between researchers and field teams, we will geographically expand the modeling techniques in this study to other cashew growing regions, and then to other major smallholder tree crops such as mango, avocado, shea, and macadamia, helping to improve the livelihoods of millions of smallholder farmers globally. In addition, based on tree crop maps, we can further understand the contribution of tree crops to carbon stocks and the benefits of expanding tree crop planting area for carbon sequestration, which could complement the research on the role of non-forest trees in mitigating climate change (Brandt et al., 2020; Mugabowindekwe et al., 2022; Skole et al., 2021).

**5 Conclusion**

In this study, we mapped the spatial distribution of cashew plantations from 2015 to 2021 and cashew planting density in 2021 with our developed tree crop mapping algorithms. Combining high-resolution Planet Basemaps and aerial imagery, even with limited ground truth labels, the STCA and U-Net showed promising performance in mapping cashew plantation locations in each of the four years. The methods and data sources used allowed us to achieve this result even in the face of difficult challenges that included heterogeneous landscapes and irregularly-planted smallholder farms, similar spectral signatures between cashew and other trees, pervasive and year-round clouds and shadows, and frequent land-use changes. We found that cashew plantation areas in Benin almost doubled since 2015 to $519 \pm 20$ kha in 2021. With the self-supervised learning model CASTC, the cashew plantation planting density map provided important information to assist in identifying regions with the greatest need for guidance on tree-spacing practices. Although the tree crop classification algorithms in this study were designed for mapping cashew plantations in Benin, they can be adapted in the future for other cashew growing regions and to map the distribution and planting density of other smallholder tree crops.

**Acknowledgements**

This research is supported by the USDA Foreign Agricultural Service and the BeninCajù Program of TechnoServe. The authors acknowledge Seth Ayim and Martin Boton for their early assistance in field ground truth collection and dissemination, Yufeng Yang for helping annotating ground truth labels, and Nanshan You for suggesting sampling strategies.

**Reference**

Alom, M.Z., Hasan, M., Yakopcic, C., Taha, T.M., & Asari, V.K., 2018. Recurrent residual convolutional neural network based on u-net (r2u-net) for medical image segmentation. arXiv preprint arXiv:1802.06955.

Atzberger, C., (2013). Advances in Remote Sensing of Agriculture: Context Description, Existing Operational Monitoring Systems and Major Information Needs. Remote Sens., 5, 949-981.

Ballester-Berman, J.D., & Rastoll-Gimenez, M., 2021. Sensitivity Analysis of Sentinel-1 Backscatter to Oil Palm Plantations at Pluriannual Scale: A Case Study in Gabon, Africa. Remote Sens., 13, 2075.

Basso, B., & Liu, L., 2019. Chapter Four - Seasonal crop yield forecast: Methods, applications, and accuracies. In D.L. Sparks (Ed.), Adv. Agron. (201-255): Academic Press.

Becker-Reshef, I., Justice, C., Barker, B., Humber, M., Rembold, F., Bonifacio, R., Zappacosta, M., Budde, M., Magadzire, T., Shitote, C., Pound, J., Constantino, A., Nakalembe, C., Mwangi, K., Sobue, S., Newby, T., Whitcraft, A., Jarvis, I., & Verdin, J., 2020. Strengthening agricultural decisions in countries at risk of food insecurity: The GEOGLAM Crop Monitor for Early Warning. Remote Sens. Environ., 237, 111553.

Brandt, M., Tucker, C.J., Kariryaa, A., Rasmussen, K., Abel, C., Small, J., Chave, J., Rasmussen, L.V., Hiernaux, P., Diouf, A.A., Kergoat, L., Mertz, O., Igel, C., Gieseke, F., Schöning, J., Li, S., Melocik, K., Meyer, J., Sinno, S., Romero, E., Glennie, E., Montagu, A., Dendoncker, M., & Fensholt, R., 2020. An unexpectedly large count of trees in the West African Sahara and Sahel. Nature, 587, 78-82.

Burke, M., & Lobell, D.B., 2017. Satellite-based assessment of yield variation and its determinants in smallholder African systems. Proc. Natl. Acad. Sci., 114, 2189-2194.

Burnett, M.W., White, T.D., McCauley, D.J., De Leo, G.A., & Micheli, F., 2019. Quantifying coconut palm extent on Pacific islands using spectral and textural analysis of very high resolution imagery. Int. J. Remote Sens., 40, 7329-7355.

Cheng, Y., Yu, L., Cracknell, A.P., & Gong, P., 2016. Oil palm mapping using Landsat and PALSAR: A case study in Malaysia. Int. J. Remote Sens., 37, 5431-5442.

Chivasa, W., Mutanga, O., & Biradar, C., 2017. Application of remote sensing in estimating maize grain yield in heterogeneous African agricultural landscapes: A review. Int. J. Remote Sens., 38, 6816-6845.

Cui, B., Huang, W., Ye, H., & Chen, Q., 2022. The Suitability of PlanetScope Imagery for Mapping Rubber Plantations. Remote Sens., 14, 1061.

Descals, A., Szantoi, Z., Meijaard, E., Sutikno, H., Rindanata, G., & Wich, S., 2019. Oil palm (Elaeis guineensis) mapping with details: smallholder versus industrial plantations and their extent in Riau, Sumatra. Remote Sens., 11, 2590.

Dong, J., Xiao, X., Chen, B., Torbick, N., Jin, C., Zhang, G., & Biradar, C., 2013. Mapping deciduous rubber plantations through integration of PALSAR and multi-temporal Landsat imagery. Remote Sens. Environ., 134, 392-402.

Dong, J., Xiao, X., Sheldon, S., Biradar, C., & Xie, G., 2012. Mapping tropical forests and rubber plantations in complex landscapes by integrating PALSAR and MODIS imagery. ISPRS J. Photogramm. Remote Sens., 74, 20-33.

Duguma, L., Minang, P., Director, A., Woldeyohanes, T., & Muthee, K., 2021 Cashew: An emerging tree commodity in African drylands for livelihoods improvement and ecosystem restoration, in: Minang PA, Duguma LA, van Noordwijk M, (Eds.), Tree Commodities and Resilient Green Economies in Africa. World Agroforestry (ICRAF), Nairobi, Kenya. https://apps.worldagroforestry.org/downloads/Publications/PDFS/BC22001.pdf (accessed 10/01/2022).

Elshamli, A., Taylor, G.W., Berg, A., & Areibi, S., 2017. Domain Adaptation Using Representation Learning for the Classification of Remote Sensing Images. IEEE J. Sel. Top. Appl. Earth Obs. Remote Sens., 10, 4198-4209.

ESA, 2017. ESA CCI LC 2016. https://2016africalandcover20m.esrin.esa.int (accessed 05/20/2022).

ESA, 2021. ESA WorldCover 10 m 2020. https://esa-worldcover.org (accessed 05/20/2022).

Gal, Y., & Ghahramani, Z., 2016. Dropout as a bayesian approximation: Representing model uncertainty in deep learning. In, international conference on machine learning (1050-1059): PMLR.

Ghosh, R., Ravirathinam, P., Jia, X., Lin, C., Jin, Z., & Kumar, V., 2021. Attention-augmented Spatio-Temporal Segmentation for Land Cover Mapping. In, 2021 IEEE International Conference on Big Data (Big Data) (1399-1408): IEEE.

Ghosh, R., Jia, X., Yin, L., Lin, C., Jin, Z., & Kumar, V., 2022. Clustering augmented self-supervised learning: an application to land cover mapping. In, Proceedings of the 30th International Conference on Advances in Geographic Information Systems (1-10).

Gomes, T., Freitas, E., Watanabe, P., Guerreiro, M., Sousa, A., & Ferreira, A., 2018. Dehydrated cashew apple meal in the feeding of growing rabbits. Semina: Ciências Agrárias, 39, 757.

Graves, A., & Schmidhuber, J., 2005. Framewise phoneme classification with bidirectional LSTM and other neural network architectures. Neural Netw., 18, 602-610.

Gutiérrez-Vélez, V.H., & DeFries, R., 2013. Annual multi-resolution detection of land cover conversion to oil palm in the Peruvian Amazon. Remote Sens. Environ., 129, 154-167.

Jain, M., Mondal, P., DeFries, R.S., Small, C., & Galford, G.L., 2013. Mapping cropping intensity of smallholder farms: A comparison of methods using multiple sensors. Remote Sens. Environ., 134, 210-223.

Ji, S., Zhang, C., Xu, A., Shi, Y., & Duan, Y., 2018. 3D convolutional neural networks for crop classification with multi-temporal remote sensing images. Remote Sens., 10, 75.

Jia, X., Khandelwal, A., Nayak, G., Gerber, J., Carlson, K., West, P., & Kumar, V., 2017. Incremental dual-memory lstm in land cover prediction. In, Proceedings of the 23rd ACM SIGKDD international conference on knowledge discovery and data mining (867-876).

Jia, X., Li, S., Khandelwal, A., Nayak, G., Karpatne, A., & Kumar, V., 2019. Spatial context-aware networks for mining temporal discriminative period in land cover detection. In, Proceedings of the 2019 SIAM International Conference on Data Mining (513-521): SIAM.

Joyce, J.M., 2011. Kullback-leibler divergence. International encyclopedia of statistical science (720-722): Springer.

Houborg, R., & McCabe, M.F., 2018. A cubesat enabled spatio-temporal enhancement method (CESTEM) utilizing Planet, Landsat and MODIS data. Remote Sens. Environ., 209, 211-226.

Hunt, D.A., Tabor, K., Hewson, J.H., Wood, M.A., Reymondin, L., Koenig, K., Schmitt-Harsh, M., & Follett, F., 2020. Review of Remote Sensing Methods to Map Coffee Production Systems. Remote Sens., 12, 2041.

Hu, Q., Sulla-Menashe, D., Xu, B., Yin, H., Tang, H., Yang, P., & Wu, W., 2019. A phenology-based spectral and temporal feature selection method for crop mapping from satellite time series. Int. J. Appl. Earth Obs. Geoinf., 80, 218-229.

Kawakubo, F.S., & Pérez Machado, R.P., 2016. Mapping coffee crops in southeastern Brazil using spectral mixture analysis and data mining classification. Int. J. Remote Sens., 37, 3414-3436.

Kumar, S., & Jayagopal, P., 2021. Delineation of field boundary from multispectral satellite images through U-Net segmentation and template matching. Ecol. Inf., 64, 101370.

Lambert, M.-J., Traoré, P.C.S., Blaes, X., Baret, P., & Defourny, P., 2018. Estimating smallholder crops production at village level from Sentinel-2 time series in Mali's cotton belt. Remote Sens. Environ., 216, 647-657.

Lin, C., Jin, Z., Mulla, D., Ghosh, R., Guan, K., Kumar, V., & Cai, Y., 2021. Toward Large-Scale Mapping of Tree Crops with High-Resolution Satellite Imagery and Deep Learning Algorithms: A Case Study of Olive Orchards in Morocco. Remote Sens., 13, 1740.

Lowder, S.K., Skoet, J., & Raney, T., 2016. The number, size, and distribution of farms, smallholder farms, and family farms worldwide. World Development, 87, 16-29.

Luong, M.-T., Pham, H., & Manning, C.D., 2015. Effective approaches to attention-based neural machine translation. arXiv preprint arXiv:1508.04025.

MacQueen, J., 1967. Classification and analysis of multivariate observations. In, 5th Berkeley Symp. Math. Statist. Probability (281-297).

MAEP-Benin, 2017. Plan Stratégique de Développement du Secteur Agricole (PSDSA) 2025 et Plan National d'Investissements Agricoles et de Sécurité Alimentaire et Nutritionnelle PNIASAN. https://ecowap.ecowas.int/media/ecowap/naip/files/BENIN_SlM6akD.pdf (accessed 10/01/2022).

MAEP-Benin, 2020. ÉTAT DE MISE EN ŒUVRE DU PAG 2016-2021. https://beninrevele.bj/doc/32/download (accessed 10/01/2022).

Maes, W.H., & Steppe, K., 2019. Perspectives for Remote Sensing with Unmanned Aerial Vehicles in Precision Agriculture. Trends Plant Sci., 24, 152-164.

Martini, M., Mazzia, V., Khaliq, A., & Chiaberge, M., 2021. Domain-Adversarial Training of Self-Attention-Based Networks for Land Cover Classification Using Multi-Temporal Sentinel-2 Satellite Imagery. Remote Sens., 13, 2564.

Maskell, G., Chemura, A., Nguyen, H., Gornott, C., & Mondal, P., 2021. Integration of Sentinel optical and radar data for mapping smallholder coffee production systems in Vietnam. Remote Sens. Environ., 266, 112709.

Masiza, W., Chirima, J.G., Hamandawana, H., & Pillay, R., 2020. Enhanced mapping of a smallholder crop farming landscape through image fusion and model stacking. Int. J. Remote Sens., 41, 8739-8756.

Mazzia, V., Khaliq, A., & Chiaberge, M., 2019. Improvement in land cover and crop classification based on temporal features learning from Sentinel-2 data using recurrent-convolutional neural network (R-CNN). Appl. Sci., 10, 238.

Mugabowindekwe, M., Brandt, M., Chave, J., Reiner, F., Skole, D., Kariryaa, A., Igel, C., Hiernaux, P., Ciais, P., Mertz, O., Tong, X., Li, S., Rwanyiziri, G., Dushimiyimana, T., Ndoli, A., Uwizeyimana, V., Lillesø, J.-P., Gieseke, F., Tucker, C., Saatchi, S., & Fensholt, R., 2022. Nation-wide mapping of tree level carbon stocks in Rwanda. In: Research Square.

Nellis, M.D., Price, K.P., & Rundquist, D., 2009. Remote sensing of cropland agriculture. The SAGE handbook of remote sensing, 1, 368-380.

Olofsson, P., Foody, G.M., Stehman, S.V. and Woodcock, C.E., 2013. Making better use of accuracy data in land change studies: Estimating accuracy and area and quantifying uncertainty using stratified estimation. Remote Sens. Environ., 129, 122-131.

Olofsson, P., Foody, G.M., Herold, M., Stehman, S.V., Woodcock, C.E. and Wulder, M.A., 2014. Good practices for estimating area and assessing accuracy of land change. Remote Sens. Environ., 148, 42-57.

Pereira, S.C., Lopes, C., & Pedro Pedroso, J., 2022. Mapping Cashew Orchards in Cantanhez National Park (Guinea-Bissau). Remote Sens. Appl.: Soc. Environ., 26, 100746.

Planet, 2022a. PlanetScope Constellation and Sensor Overview. https://developers.planet.com/docs/data/planetscope/ (accessed 05/20/2022).

Planet, 2022b. Basemaps Overview. https://developers.planet.com/docs/basemaps/ (accessed 05/20/2022).

Planet, 2022c. SuperDove's 8 Spectral Bands. https://content.planet.com/c/65?x=XdDOks&lx=vlHTzf&mkt_tok=OTk3LUNISC0yNjUAAAGEChYFMnt1TyiacP7mjfeFdPERCL_4teyoha44Y0aoY51Li5SymeU6vPJ4J2zZVRr3Q051pMo2dd9GRSUB68HJhjSDpQmduwQ3EggMLRkh2A (accessed 05/20/2022).

Planet, 2022d. Next Generation PlanetScope. https://www.planet.com/pulse/planet-launches-next-generation-planetscope-with-eight-spectral-bands-and-quality-improvements (accessed 05/20/2022).

PNIASAN-Benin, 2022. Benin/Regulatory framework for public and private investments: Development of PNIASAN 3rd generation launched. https://www.jumelages-partenariats.com/en/actualites.php?n=16949 (accessed 10/01/2022).

Rao, P., Zhou, W., Bhattarai, N., Srivastava, A.K., Singh, B., Poonia, S., Lobell, D.B., & Jain, M., 2021. Using Sentinel-1, Sentinel-2, and Planet imagery to map crop type of smallholder farms. Remote Sens, 13, 1870.

Roy, D.P., Huang, H., Houborg, R., & Martins, V.S., 2021. A global analysis of the temporal availability of PlanetScope high spatial resolution multi-spectral imagery. Remote Sens. Environ., 264, 112586.

Ronneberger, O., Fischer, P., & Brox, T., 2015. U-net: Convolutional networks for biomedical image segmentation. In, International Conference on Medical image computing and computer-assisted intervention (234-241): Springer.

Samberg, L.H., Gerber, J.S., Ramankutty, N., Herrero, M., & West, P.C., 2016. Subnational distribution of average farm size and smallholder contributions to global food production. Environ. Res. Lett., 11, 124010.

Sanjeeva, S.K., Pinto, M.P., Narayanan, M.M., Kini, G.M., Nair, C.B., SubbaRao, P., Pullela, P.K., Ramamoorthy, S., & Barrow, C.J., 2014. Distilled technical cashew nut shell liquid (DT-CNSL) as an effective biofuel and additive to stabilize triglyceride biofuels in diesel. Renewable energy, 71, 81-88.

Schneibel, A., Stellmes, M., Röder, A., Frantz, D., Kowalski, B., Haß, E., & Hill, J., 2017. Assessment of Spatio-temporal changes of smallholder cultivation patterns in the Angolan Miombo belt using segmentation of Landsat time series. Remote Sens. Environ., 195, 118-129.

Shi, X., Chen, Z., Wang, H., Yeung, D.-Y., Wong, W.-K., & Woo, W.-c., 2015. Convolutional LSTM network: A machine learning approach for precipitation nowcasting. Advances in neural information processing systems, 28.

Singh, M., Evans, D., Chevance, J.-B., Tan, B.S., Wiggins, N., Kong, L., & Sakhoeun, S., 2018. Evaluating the ability of community-protected forests in Cambodia to prevent deforestation and degradation using temporal remote sensing data. Ecol. Evol., 8, 10175-10191.

Sishodia, R.P., Ray, R.L., & Singh, S.K., 2020. Applications of remote sensing in precision agriculture: A review. Remote Sens., 12, 3136.

Skole, D.L., Mbow, C., Mugabowindekwe, M., Brandt, M.S., & Samek, J.H., 2021. Trees outside of forests as natural climate solutions. Nat. Clim. Change, 11, 1013-1016.

Somers, B., & Asner, G.P., 2013. Multi-temporal hyperspectral mixture analysis and feature selection for invasive species mapping in rainforests. Remote Sens. Environ., 136, 14-27.

Stehman, S.V., 2014. Estimating area and map accuracy for stratified random sampling when the strata are different from the map classes. Int. J. Remote Sens., 35, 4923-4939

Tridawati, A., Wikantika, K., Susantoro, T.M., Harto, A.B., Darmawan, S., Yayusman, L.F., & Ghazali, M.F., 2020. Mapping the distribution of coffee plantations from multi-resolution, multi-temporal, and multi-sensor data using a random forest algorithm. Remote Sens., 12, 3933.

Tuia, D., Persello, C., & Bruzzone, L., 2016. Domain adaptation for the classification of remote sensing data: An overview of recent advances. IEEE Geosci. Remote Sens. Mag., 4, 41-57.

UN, 2015. Sustainable Development Goals. https://www.un.org/sustainabledevelopment (accessed 05/20/2022).

UNCTAD, 2021. Commodities at a Glance: Special issue on cashew nuts. https://unctad.org/system/files/official-document/ditccom2020d1_en.pdf (accessed 05/20/2022).

Van der Maaten, L., & Hinton, G., 2008. Visualizing data using t-SNE. Journal of machine learning research, 9.

Vincenzi, S., Porrello, A., Buzzega, P., Cipriano, M., Fronte, P., Cuccu, R., Ippoliti, C., Conte, A., & Calderara, S., 2021. The color out of space: learning self-supervised representations for earth observation imagery. In, 2020 25th International Conference on Pattern Recognition (ICPR) (3034-3041): IEEE.

Waarts, Y., Janssen, V., Aryeetey, R., Onduru, D., Heriyanto, D., Aprillya, S., N'Guessan, A., Courbois, L., Bakker, D., & Ingram, V., 2021. Multiple pathways towards achieving a living income for different types of smallholder tree-crop commodity farmers. Food Security, 13, 1467-1496.

Wei, S., Zhang, H., Wang, C., Wang, Y., & Xu, L., 2019. Multi-Temporal SAR Data Large-Scale Crop Mapping Based on U-Net Model. Remote Sens., 11, 68.

Xu, Y., Yu, L., Ciais, P., Li, W., Santoro, M., Yang, H., & Gong, P., 2022. Recent expansion of oil palm plantations into carbon-rich forests. Nat. Sustainability, 5, 574-577.

Yin, L., You, N., Zhang, G., Huang, J., & Dong, J., 2020. Optimizing feature selection of individual crop types for improved crop mapping. Remote Sens., 12, 162.

Zeng, Y., Hao, D., Huete, A., Dechant, B., Berry, J., Chen, J.M., Joiner, J., Frankenberg, C., Bond-Lamberty, B., Ryu, Y., Xiao, J., Asrar, G.R., & Chen, M., 2022. Optical vegetation indices for monitoring terrestrial ecosystems globally. Nat. Rev. Earth Environ., 3, 477-493.

Zou, K., Chen, X., Zhang, F., Zhou, H., & Zhang, C., 2021. A Field Weed Density Evaluation Method Based on UAV Imaging and Modified U-Net. Remote Sens., 13, 310.